\documentclass{article}
\usepackage{ijcai21}

\usepackage{times}
\usepackage{soul}
\usepackage{url}
\usepackage[hidelinks]{hyperref}
\usepackage[utf8]{inputenc}
\usepackage[small]{caption}
\usepackage{graphicx}
\usepackage{amsmath}
\usepackage{amsthm}
\usepackage{amsfonts}
\usepackage{amssymb}
\usepackage{booktabs}
\usepackage{multicol}
\usepackage{multirow}
\usepackage{algorithm}
\usepackage{algorithmic}
\usepackage{textcomp}

\usepackage{bm}
\usepackage{booktabs}

\usepackage{arydshln}
\usepackage{tikz}
\usepackage{pgfplots}
\usepackage{subcaption}
\usepackage{graphics}
\usepackage{adjustbox}
\usepackage{comment}
\usepackage{dsfont}

\newcommand{\regu}{\mathrm{reg}}
\newcommand{\aug}{\mathrm{aug}}
\newcommand{\ins}{\mathrm{ins}}
\newcommand{\del}{\mathrm{del}}
\newcommand{\rep}{\mathrm{rep}}
\newcommand{\cor}{\mathrm{cor}}
\newcommand{\incor}{\mathrm{incor}}
\newcommand{\tot}{\mathrm{tot}}

\newcommand{\ori}{\mathrm{ori}}
\mathchardef\shyp="2D 

\newcommand{\auc}[2]{#1 $\pm$ {\tiny #2}}
\newcommand{\aucgain}[3]{#1 $\pm$ {\tiny #2}}
\newcommand{\boldauc}[2]{\textbf{#1} $\pm$ {\tiny #2}}
\newcommand{\boldaucgain}[3]{\textbf{#1} $\pm$ {\tiny #2}}
\newcommand{\upar}{$\uparrow$}
\newcommand{\doar}{$\downarrow$}

\title{Consistency and Monotonicity Regularization for Neural Knowledge Tracing}

\author{
Seewoo Lee$^1$\and
Youngduck Choi$^1$\and
Juneyoung Park$^1$\footnote{Contact Author}\and
Byungsoo Kim$^1$\And
Jinwoo Shin$^2$\\
\affiliations
$^1$Riiid! AI Research\\
$^2$Graduate School of AI, KAIST\\
\emails
\{seewoo.lee, youngduck.choi, juneyoung.park, byungsoo.kim\}@riiid.co,
jinwoos@kaist.ac.kr
}
\begin{document}

\maketitle

\begin{abstract}
Knowledge Tracing (KT), tracking a human's knowledge acquisition, is a central component in online learning and AI in Education.
In this paper, we present a simple, yet effective strategy to improve the generalization ability of KT models:
we propose three types of novel data augmentation, coined replacement, insertion, and deletion, along with corresponding regularization losses that impose certain consistency or monotonicity biases on model's predictions for the original and augmented sequence. 
Extensive experiments on various KT benchmarks show that our regularization scheme consistently improves the model performances, under 3 widely-used neural networks and 4 public benchmarks, e.g.,
it yields 6.3\% improvement in AUC under the DKT model and the ASSISTmentsChall dataset.
\end{abstract}

% \vspace{-0.1in}
\section{Introduction}

% covid-19, aied
In recent years, Artificial Intelligence in Education (AIEd) has gained much attention as an emerging field to elevate educational technology. 
Especially due to the circumstances from the COVID-19 pandemic, much of the education industry was forcibly moved to an online environment which inevitably allowed much opportunity to utilize educational data. The ability to diagnose students through data and provide personalized learning paths have become a critical edge to uplift online education. The most fundamental aspect of assessing student's current knowledge state has been the focus of AIED research, a task commonly known as Knowledge Tracing(KT). Creating a more precise and a robust KT model has become an essential path to develop a highly effective and AI-based educational system.
%,
%which has generated tremendous amount of student interaction data. 
%As a result,  AIEd has become more prominent because of its ability to diagnose students automatically and provide personalized learning paths. 
%High-quality diagnosis and educational content recommendation require good understanding of students’ current knowledge state, and it is essential to model %their learning behavior precisely. 
%Due to this, Knowledge Tracing (KT), a task of modeling a student’s evolution 
%of knowledge over time, has become one of the most central tasks in AIEd research.

\begin{figure}[h]
    \centering
    \begin{subfigure}[b]{0.2\textwidth}
        \centering
        \includegraphics[width=\textwidth]{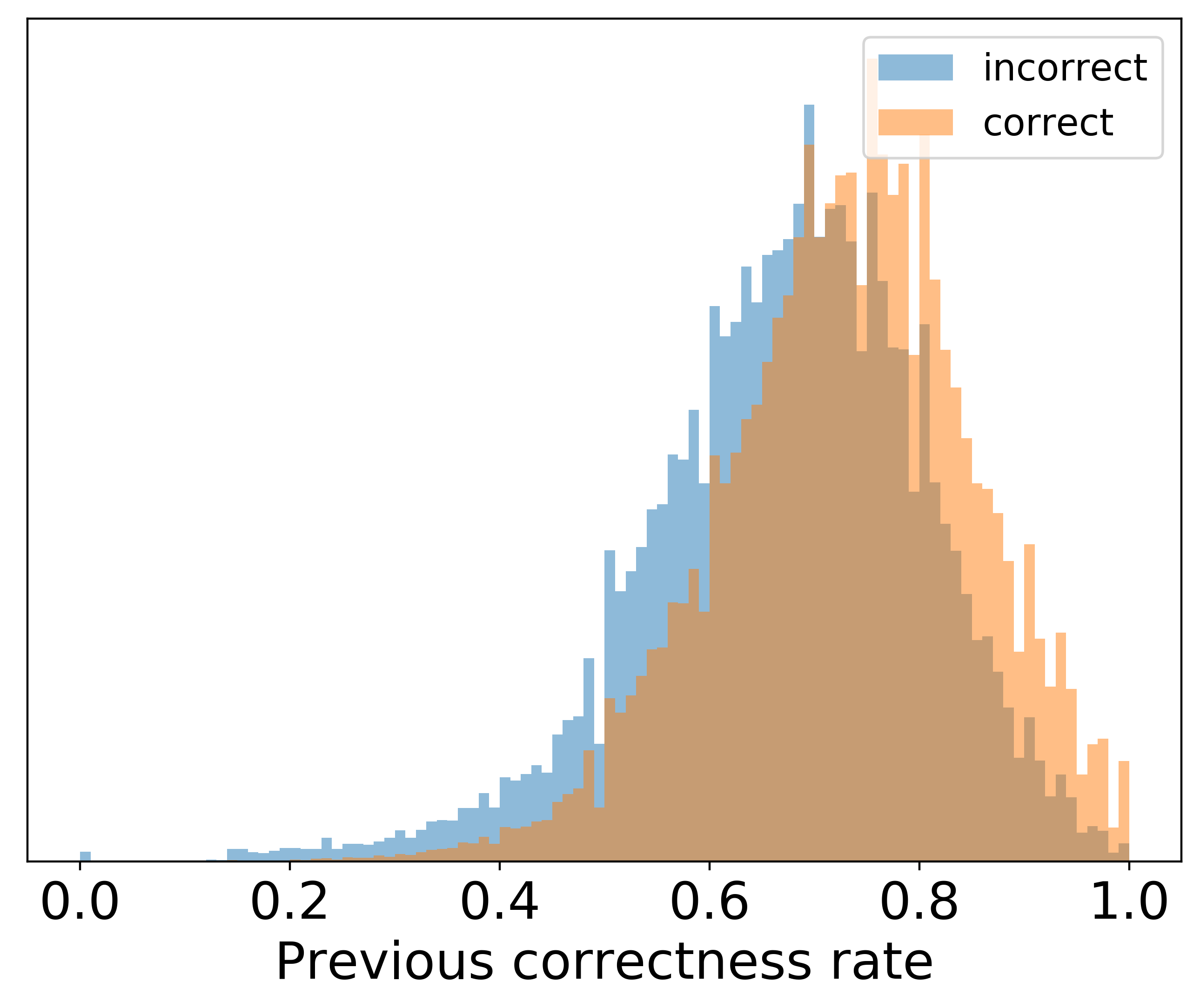}
        \caption[Network2]%
        {{\small ASSISTments2015}}    
        \label{fig:mean and std of net14}
    \end{subfigure}
    \hfill
    \begin{subfigure}[b]{0.2\textwidth}  
        \centering 
        \includegraphics[width=\textwidth]{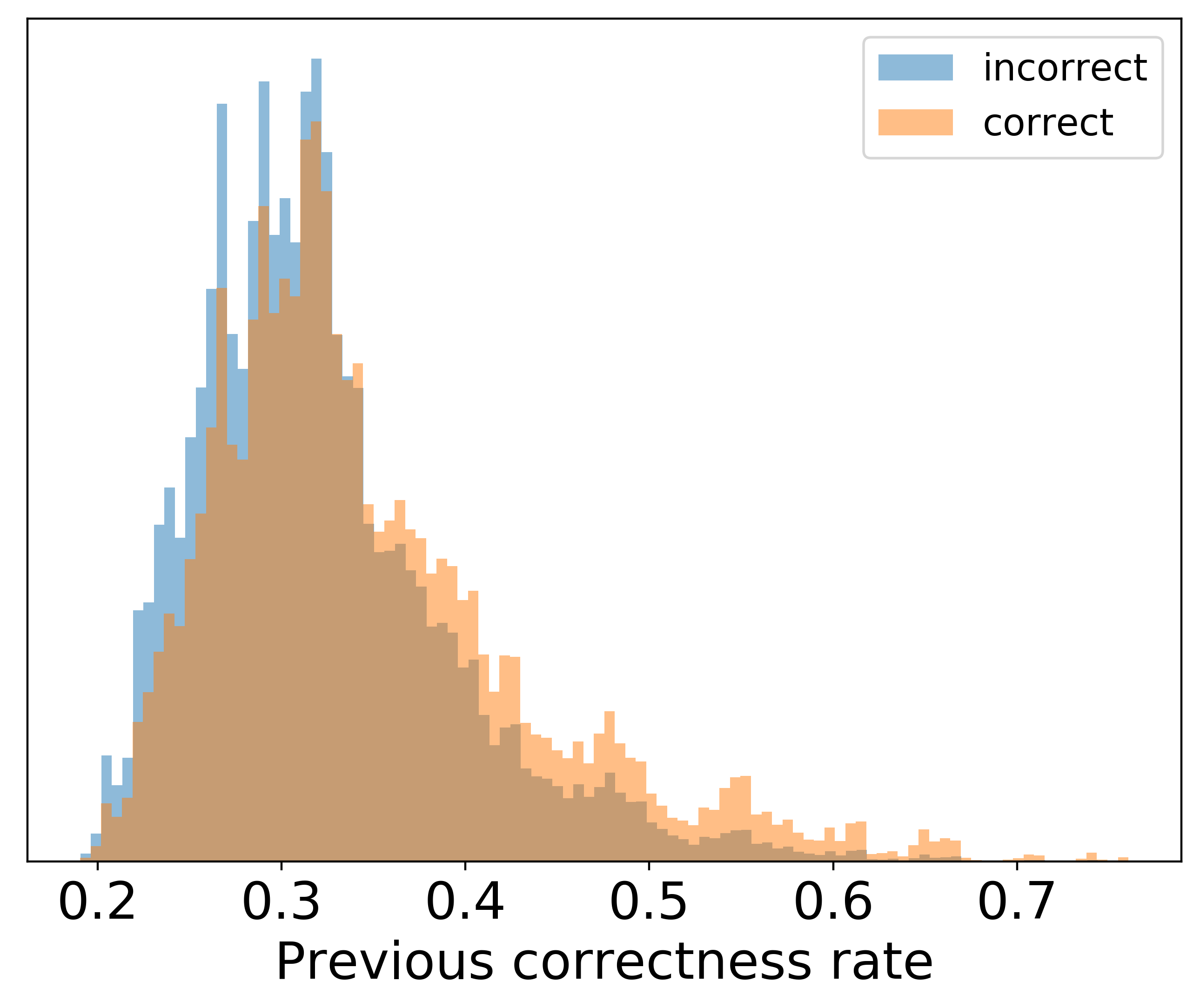}
        \caption[]%
        {{\small ASSISTmentsChall}}    
        \label{fig:mean and std of net24}
    \end{subfigure}
    \vskip\baselineskip
    \begin{subfigure}[b]{0.2\textwidth}   
        \centering 
        \includegraphics[width=\textwidth]{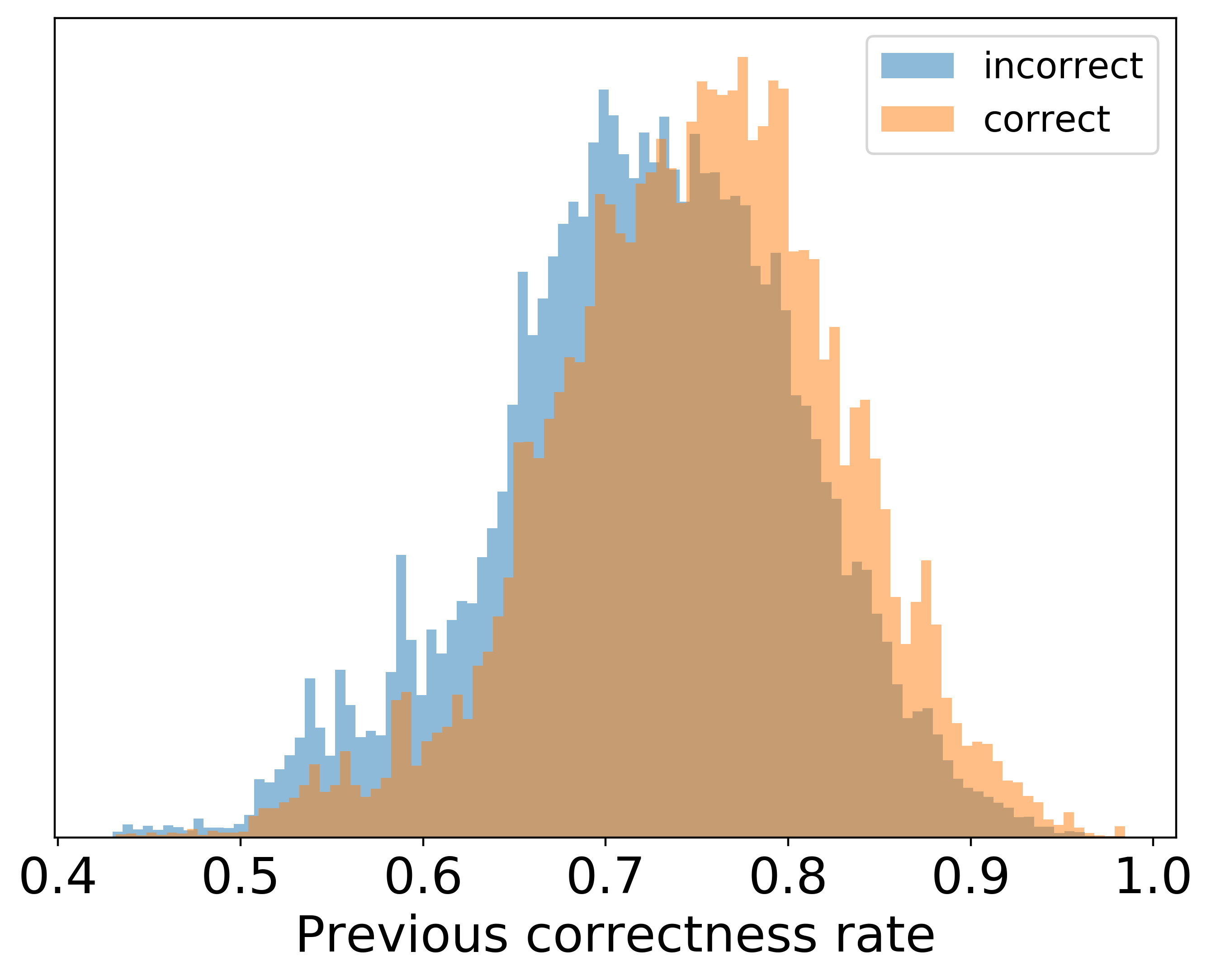}
        \caption[]%
        {{\small STATICS2011}}    
        \label{fig:mean and std of net34}
    \end{subfigure}
    \hfill
    \begin{subfigure}[b]{0.2\textwidth}   
        \centering 
        \includegraphics[width=\textwidth]{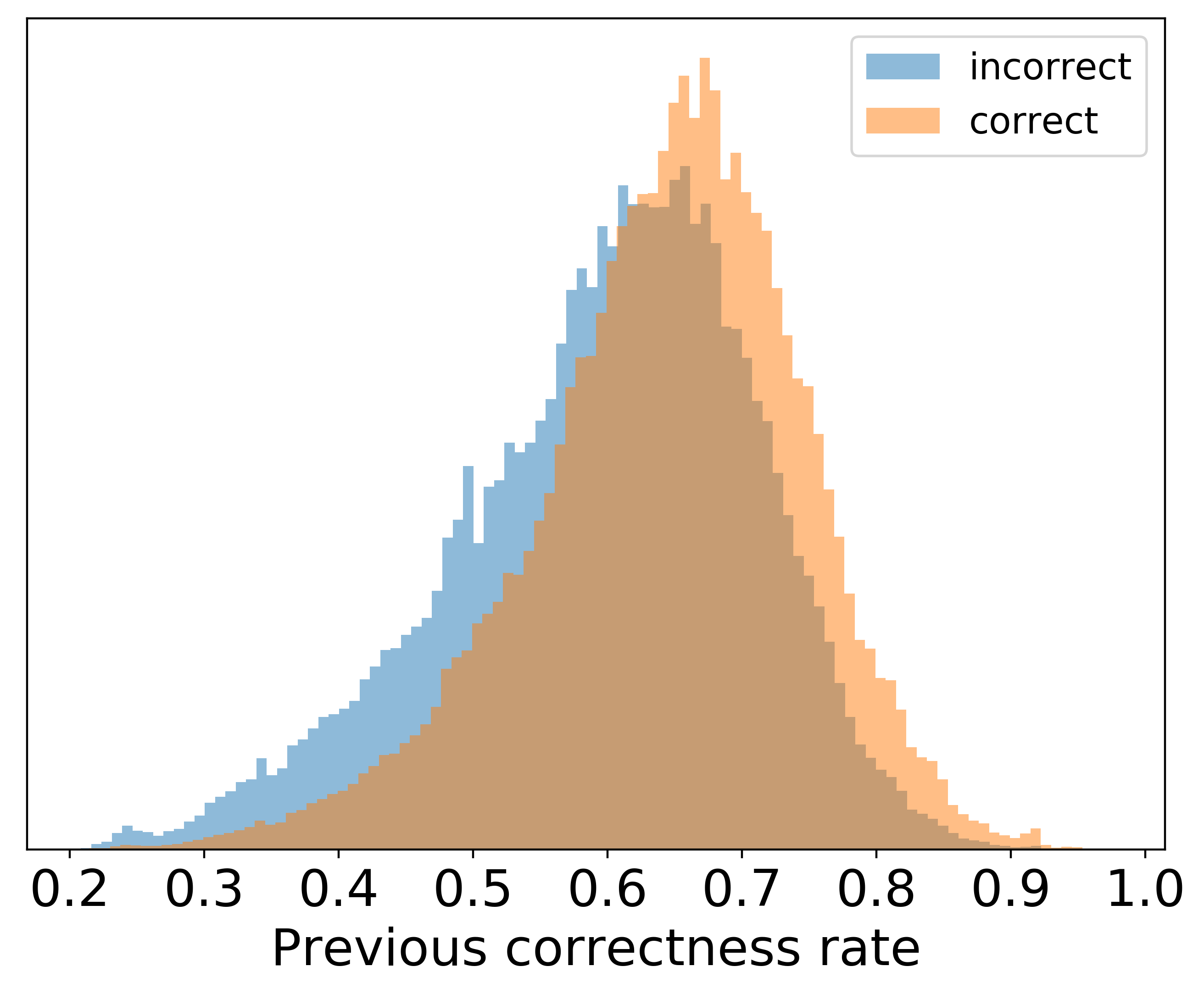}
        \caption[]%
        {{\small EdNet-KT1}}    
        \label{fig:mean and std of net44}
    \end{subfigure}
    \caption[ The average and standard deviation of critical parameters ]
    {
    \small Distribution of the correctness rate of past interactions when the response correctness of current interaction is fixed, for 4 knowledge tracing benchmark datasets. 
    Orange (resp. blue) represents the distribution of correctness rate (of past interactions) where current interaction's response is correct (resp. incorrect). $x$ axis represents previous interactions' correctness rates (values in $[0, 1]$). The orange distribution lean more to the right than the blue distribution, which shows the monotonicity nature of the interaction datasets.
    % \textcolor{blue}{See Section \ref{sec:mono_data} for details.}} 
    }
    \label{fig:data_monotonicity}
    \vspace{-0.20in}
\end{figure}

% KT

Since the work of %Deep Knowledge Tracing 
\cite{piech2015deep}, deep neural networks have been widely used for the KT modeling. 
Current research trends in the KT literature concentrate on building more sophisticated, complex and large-scale models, inspired by model architectures from Natural Language Processing (NLP), such as 
LSTM \cite{hochreiter1997long}
or Transformer \cite{vaswani2017attention} architectures. 
% Further investigations to introduce additional components based on educational contexts such as  textual information  or students' forgetting behaviors \cite{huang2019ekt,ghosh2020context}. 
Further investigations to introduce additional components based on educational contexts such as  textual information \cite{huang2019ekt}. 
However, not all educational data are sufficiently large and more often than not, the larger model sizes lead to overfitting and ultimately hurt the model's generalizabiliy \cite{gervet2020deep} (See Figure 1 of the Appendix).
To the best of our knowledge, only a handful of the literature addresses such issues and even then, the scope is limited to regularization \cite{yeung2018addressing,sonkar2020qdkt}.

\begin{figure}
% \vspace{-0.1in}
    \centering
        \includegraphics[width=3.3in]{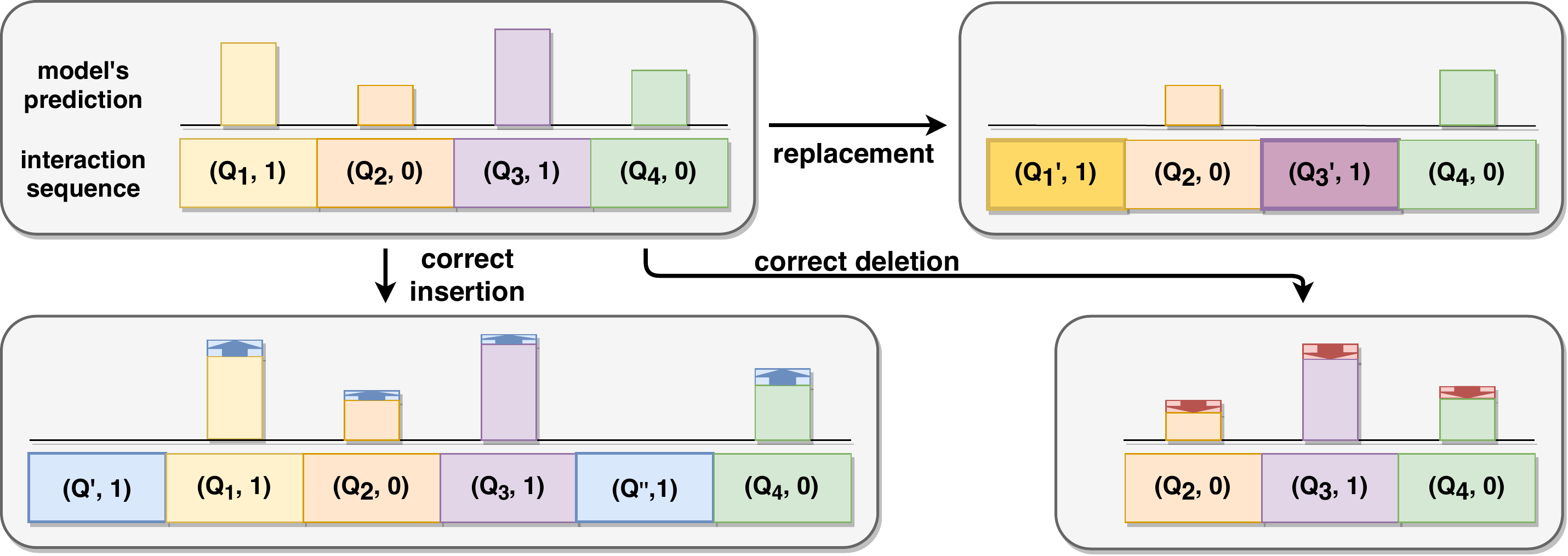}
    \caption{
    Augmentation strategies and corresponding bias on model's predictions. 
    Each tuple represents question id and response of the student's interaction (1 means correct). Replacing interactions with similar questions ($Q_{1}, Q_{3}$ to $Q_{1}', Q_{3}'$)  does not change model's predictions drastically. Introducing new interactions with correct responses $(Q', Q'')$ increases model's predictions, but deleting such interaction $(Q_1, 1)$ decreases them.}
    \label{fig:aug_example}
    \vspace{-0.2in}
\end{figure}

To address the issue, we propose simple, yet effective data augmentation strategies for improving the generalization ability of KT models, along with novel regularization losses for each strategy. 
In particular, we suggest three types of data augmentation, coined (skill-based) replacement, insertion, and deletion.
Specifically, we generate augmented (training) samples by randomly replacing questions that a student solved with similar questions or inserting/deleting interactions with fixed responses. 
Furthermore, during training, we impose certain consistency (for replacement) and monotonicity (for insertion/deletion) bias on the model's predictions by optimizing corresponding regularization losses that compares the original and the augmented interaction sequences.  Such regularization strategies are motivated from our observation that existing knowledge tracing models' prediction often fails to satisfy the consistency and monotonicity condition, e.g., see Figure \ref{fig:heatmap} in Section \ref{sec:exp}.
% forces models to satisfy consistency and monotonicity constraints by including \emph{consistency} and \emph{monotonicity} losses in a training objective. 
% \textcolor{red}{JW: Provide a bit more detailed intuition on our proposed components and why it should work. Otherwise, readers do have no clue how skill-based replacement, insertion, and deletion (and corresponding regularization losses) look like. This paragraph should be of at least 10 lines or longer}
Here, our intuition behind the proposed consistency regularization %with skill-based replacement 
is that the model's output for two interaction sequences with same response logs for similar questions should be close. 
Next, the proposed monotonicity regularization is designed to enforce the model's prediction to be monotone with respect to the number of questions that are correctly (or incorrectly) answered, i.e., a student is more likely to be correct (or incorrectly) in the next question if the student was more correct in the past.
By analyzing the distribution of previous correctness rate, we can observe that the existing student interaction is indeed monotonic as shown in Figure \ref{fig:data_monotonicity}.
% and Section \ref{sec:mono_data} for details.
The overall augmentation and regularization strategies are sketched in Figure \ref{fig:aug_example}.

We demonstrate the effectiveness of the proposed method with 3 widely used neural knowledge tracing models - DKT \cite{piech2015deep}, DKVMN \cite{zhang2017dynamic}, and SAINT \cite{choi2020towards} -  on 4 public benchmark datasets - ASSISTments2015, ASSISTmentsChall, STATICS2011, and EdNet-KT1.  
Extensive experiments show that, regardless of dataset or model architecture, our scheme remarkably increases the prediction performance - 6.3\% gain in Area Under Curve (AUC) for DKT on the ASSISTmentsChall dataset.
% \textcolor{red}{JW: Provide more highlights and details on experimental results. This paragraph should be of at least 10 lines or longer.}
In particular, ours is much more effective under smaller datasets: by using only $25\%$ of the ASSISTmentsChall dataset, we improve AUC of the DKT model from 69.68\% to 75.44\%, which even surpasses the baseline performance 74.4\% with the full training set. 
We further provide various ablation studies for the selected design choices, e.g.,
%on the augmentations established the importance of the direction of monotonicity constraints and utilizing skill information in replacement. 
AUC of the DKT model on the ASSISTments2015 dataset dropped from 72.44\% to 66.48\% when we impose `reversed' (wrong) monotonicity regularization.
 The findings from the current study contribute to existing KT literature by providing a novel generalization mechanism which provide a strong baseline for future augmentation and regularization research.

% \vspace{-0.05in}
\subsection{Related Works and Preliminaries}

\textbf{Data augmentation}
%The data augmentation 
is arguably the most trustworthy technique to prevent overfitting or improve the generalizability of machine learning models.
In particular, it has been developed as an effective way to impose a domain-specific, inductive bias to a model.
For example, for computer vision models, simple image warpings such as flip, rotation, distortion, color shifting, blur, and random erasing are 
the most popular data augmentation methods \cite{shorten2019survey}. 
% More advanced techniques, e.g., augmenting images by interpolation \cite{zhang2017mixup} or by using generative adversarial networks \cite{huang2018auggan}, have been also investigated. 
For NLP models, it is popular to augment texts by replacing words with synonyms \cite{zhang2015character} or words with similar (contextualized) embeddings \cite{kobayashi2018contextual}. 
% As an alternative method, back translation \cite{yu2018qanet} generates an augmented sentence by translating a given sentence into a different language domain and translate it back to the original domain with machine translation models. 
Recently, \cite{wei2019eda} show that even simple methods like random insertion/swap/deletion could improve text classification performances. 
% In the area of speech recognition, vocal tract length normalization \cite{jaitly2013vocal}, synthesizing noisy audio \cite{hannun2014deep}, perturbing speed \cite{ko2015audio}, and augmenting spectrogram \cite{park2019specaugment} are popular as data augmentation methods.

The aforementioned data augmentation techniques have been used not only for standard supervised learning setups, but also for various unsupervised and semi-supervised learning frameworks, by imposing certain inductive biases to models. 
For example, consistency learning
\cite{berthelot2019mixmatch}
% \cite{sajjadi2016regularization,xie2019unsupervised,berthelot2019mixmatch} 
impose a consistency bias to a model so that the model's output is invariant under the augmentations, by means of training the model with consistency regularization loss (e.g. $L^{2}$-loss between outputs). 
\cite{abu1992method} suggested general theory for imposing such inductive biases (which are stated as \emph{hints}) via additional regularization losses. 
Their successes highlight the importance of domain specific knowledge for designing appropriate data augmentation strategies, but such results are rare in the domain of AIEd, especially for Knowledge Tracing.

\textbf{Knowledge tracing} (KT) is the task of modeling student knowledge over time based on the student's learning history.
Formally, for a given student interaction sequence $(I_1, \dots, I_T)$, where each  $I_t = (Q_t, R_t)$ is a pair of question id $Q_t$ and the student's response correctness $R_t \in \{0, 1\}$ ($1$ means correct), KT aims to estimate the following probability
\begin{align}
    \mathbb{P}[R_{t} =1  | I_1, I_2, \dots, I_{t-1}, Q_{t}],
\end{align}
i.e., the probability that the student answers correctly to the question $Q_t$ at $t$-th step. 
\cite{corbett1994knowledge} proposed Bayesian Knowledge Tracing (BKT) that models a student’s knowledge as a latent variable in a Hidden Markov Model. 
Also, various seq2seq architectures including LSTM \cite{hochreiter1997long}, MANN \cite{graves2016hybrid}, and Transformer \cite{vaswani2017attention} are used in the context of KT and showed their efficacy. 
Deep Knowledge Tracing (DKT) is the first deep learning based model that models student’s knowledge states as LSTM’s hidden state vectors \cite{piech2015deep}. 
Dynamic Key-Value Memory Network and its variation can exploit relationships between questions/skills with concept vectors and concept-state vectors with key and value matrices, which is more interpretable than DKT \cite{zhang2017dynamic}. 
Transformer based models
\cite{pandey2019self,choi2020towards}
% \cite{pandey2019self,choi2020towards,ghosh2020context} 
are able to learn long-range dependencies with their self-attention mechanisms and be trained in parallel. 
Utilizing additional features of interactions, such as question texts \cite{huang2019ekt} 
%, prerequisite relations \cite{chen2018prerequisite} 
and time information \cite{nagatani2019augmenting}, is another way to improve performances. 
\section{Consistency and Monotonicity Regularization for KT}
%Knowledge Tracing}

For a given set of data augmentations $\mathcal{A}$, we train KT models with the following loss:
\begin{align}
    \mathcal{L}_{\tot} = \mathcal{L}_{\ori} + \sum_{\aug \in \mathcal{A}} ( \lambda_{\aug}\mathcal{L}_{\aug} + \lambda_{\regu\shyp\aug}\mathcal{L}_{\regu\shyp\aug}),
\label{eqn:total_loss}
\end{align}
where $\mathcal{L}_{\ori}$ is the commonly used binary cross-entropy (BCE) loss for original training sequences
% for KT
% $$
% \mathcal{L}_{\KT} =  -\frac{1}{T}\sum_{1 \leq t \leq T}\ R_{t} \log p_{t} + (1 - R_{t}) \log (1 - p_{t}),
% $$
% $$
% \mathcal{L}_{\ori} = \mathbb{E}_{t}\left[-R_{t} \log p_{t} - (1 - R_{t}) \log (1 - p_{t})\right],
% \label{eqn:ori_loss}
% $$
and $\mathcal{L}_{\aug}$ are the same BCE losses for augmented sequences generated by applying augmentation strategies $\aug \in \mathcal{A}$.\footnote{For replacement and insertion, we do not include outputs for augmented interactions in  $\mathcal{L}_{\aug}$. }
$\mathcal{L}_{\regu\shyp\aug}$ are the regularization losses that impose consistency and monotonicity bias on the model's predictions for the original and augmented sequence, which will be defined in the following sections. 
% described in the previous subsections, 
% $\mathcal{A} \subseteq \{\rep, \cor\_\ins, \incor\_\ins, \cor\_\del, \incor\_\del\}$ is a set of augmentations we use while training. 
%For the replacement and insertion, we do not include newly generated interaction's output in the BCE losses for augmented sequences ($\mathcal{L}_{\KT}^{\rep}$, $\mathcal{L}_{\KT}^{\cor\_\ins}$ and $\mathcal{L}_{\KT}^{\incor\_\ins}$).
Finally, $\lambda_{\aug},\lambda_{\regu\shyp\aug}>0$ are hyperparameters to control the trade-off among $\mathcal{L}_{\ori}$, $\mathcal{L}_{\aug}$, and $\mathcal{L}_{\regu\shyp\aug}$.
%($\mu_{\aug} \in \{0, 1\}$ is actually a flag that determines whether we include the KT loss for augmented sequences in the total loss or not.)

In the following sections, we introduce the three augmentation strategies,
replacement, insertion and deletion with corresponding consistency and monotonicity regularization losses, $\mathcal{L}_{\regu\shyp\rep}$, $\mathcal{L}_{\regu\shyp\cor\_\ins}$ (or $\mathcal{L}_{\regu\shyp\incor\_\ins}$) and $\mathcal{L}_{\regu\shyp\cor\_\del}$ (or $\mathcal{L}_{\regu\shyp\incor\_\del}$).

\vspace{-0.05in}
\subsection{Replacement}

% \begin{figure}
%     \centering
%     \includegraphics[width=\textwidth]{replacement_real_last.pdf}
%     \caption{Replacement.
%     The 1st and 3rd interactions in the original sequence are replaced.
%     The predicted correctness probabilities for unreplaced questions remain the same.}
%     \label{fig:replacement}
% \end{figure}

Replacement, similar to the synonym replacement strategy in NLP, is an augmentation strategy that replaces questions in the original interaction sequence with other similar questions \emph{without changing their responses}, where \emph{similar questions} are defined as the questions that share the same skills as the original question.
Our hypothesis is that the predicted correctness probabilities for questions in an augmented interaction sequence should not change drastically compared to the original interaction sequence.
Formally, for each interaction in the original interaction sequence $(I_1, \dots, I_T)$, we randomly decide whether the interaction will be replaced or not, following the Bernoulli distribution with the probability $\alpha_{\rep}$.
If an interaction $I_{t} = (Q_{t}, R_{t})$ with a set of skills $S_{t}$ associated with the question $Q_{t}$ is set to be replaced, we determine $I_{t}^{\rep} = (Q_{t}^{\rep}, R_{t})$ by selecting a question $Q_{t}^{\rep}$ with its associated set of skills $S_{t}^{\rep}$ that satisfies $S_{t} \cap S_{t}^{\rep} \neq \emptyset$.
%\begin{enumerate}
%    \item For a set of skills $S_{t}$ associated with the question $Q_{t}$, we select a question $Q_{t}^{\rep}$ with its associated set of skills $S_{t}^{\rep}$ that $S_{t} \cap S_{t}^{\rep} \neq \emptyset$, and leave $R_{t}^{\rep}$ same as $R_{t}$.
%    \item We select a question $Q_{t}^{\rep}$ by following the same strategy in 1.
%    However, for $R_{t}^{\rep}$, we randomly determine its value either one or zero.
%    \item We randomly select a question $Q_{t}^{\rep}$ from the question pool, and leave $R_{t}^{\rep}$ same as $R_{t}$.
%    \item We select a question $Q_{t}^{\rep}$ by following the same strategy in 3.
%    However, for $R_{t}^{\rep}$, we randomly determine its value either one or zero.
%\end{enumerate}
The resulting augmented sequence $(I_{1}^{\rep}, \dots, I_{T}^{\rep})$ is generated by replacing $I_{t}$ with $I_{t}^{\rep}$ for $t \in \mathbf{R} \subset [T] = \{1,2,\dots, T\}$, where $\mathbf{R}$ is a set of indices to replace.
Then we consider the following consistency regularization loss:
\begin{align}
    % \mathcal{L}_{\rep}^{\mathrm{nro}} = \frac{1}{T - |\mathbf{R}|}\sum_{t \not\in \mathbf{R}} (p_{t} - p_{t}^{\rep})^{2}
    \mathcal{L}_{\regu\shyp\rep} = \mathbb{E}_{t\not\in \mathbf{R}}[(p_{t} - p_{t}^{\rep})^{2}],
    \label{eqn:repnro}
\end{align}
%\begin{align}
%    \mathcal{L}_{\rep}^{\mathrm{nro}} = \frac{1}{T - |\mathbf{R}|}\sum_{t \not\in \mathbf{R}} (p_{t} - p_{t}^{\rep})^{2}, \quad \mathcal{L}_{\rep}^{\mathrm{ro}} = \frac{1}{|\mathbf{R}|}\sum_{t \in \mathbf{R}} (p_{t} - p_{t}^{\rep})^{2}, \quad \mathcal{L}_{\rep}^{\mathrm{all}} = \frac{1}{T}\sum_{t\in [T]} (p_{t} - p_{t}^{\rep})^{2} \label{eqn:repall}
%\end{align}
where $p_t$ and $p_{t}^{\rep}$ are model's predicted correctness probabilities for $t$-th question of the original and augmented sequences.
The output for the replaced interactions are not included in the loss computation. 
% , and $\mathrm{nro}$ stands for \emph{not replaced only} which only considers losses for interactions that are not replaced.

Also, the replacement strategy has several variants depending on the dataset.
For instance, randomly selecting a question for $Q_{t}^{\rep}$ from the question pool is an alternative strategy when the related skill set information is not available.
It is also possible to only consider outputs for interactions that are replaced or consider outputs for all interactions in the augmented sequence for the loss computation.
We investigate the effectiveness of each strategy in Section \ref{sec:exp}.

\vspace{-0.05in}
\subsection{Insertion}

% \begin{figure}
%     \centering
%     \includegraphics[width=\textwidth]{insertion_all (1).pdf}
%     \caption{Insertion. Two new interactions are inserted in the 1st and 4th positions.
%     The predicted correctness probabilities for questions in the original interaction sequence increase (resp. decrease) as we insert new interactions with correct (resp. incorrect) responses.}
%     \label{fig:insertion}
% \end{figure}

Insertion strategy is based on the notion of a student's knowledge to be higher when more questions are answered correctly. 
Specifically, data is augmented in a \emph{monotonic} manner by inserting new interactions into the original interaction sequence.
Formally, we generate an augmented interaction sequence $(I_{1}^{\ins}, \dots, I_{T'}^{\ins})$ by inserting a correctly (resp. incorrectly) answered interaction $I_{t}^{\ins} = (Q_{t}^{\ins}, 1)$ (resp. $I_{t}^{\ins} = (Q_{t}^{\ins}, 0)$) into the original interaction sequence $(I_{1}, \dots, I_{T})$ for $t \in \mathbf{I} \subset [T']$, where the question $Q_{t}^{\ins}$ is randomly selected from the question pool and $\mathbf{I}$ with the size $\alpha_{\ins}$ proportion of the original sequence is a set of indices of inserted interactions.
Then our hypothesis is formulated as $p_{t} \leq p_{\sigma(t)}^{\ins}$ (resp. $p_{t} \geq p_{\sigma(t)}^{\ins}$), where $p_{t}$ and $p_{t}^{\ins}$ are model's predicted correctness probabilities for $t$-th question of the original and augmented sequences, respectively.
Here, $\sigma: [T] \to [T'] - \mathbf{I}$ is the order-preserving bijection which satisfies $I_{t} = I^{\ins}_{\sigma(t)}$ for $1 \leq t \leq T$. 
(For instance, in Figure \ref{fig:aug_example}, $\sigma$ sends $\{1, 2, 3, 4\}$ to $\{2, 3,4, 6\}$)
We impose our hypothesis through the following losses:
\begin{align}
    % \mathcal{L}_{\cor\_\ins} = \frac{1}{T}\sum_{ t \in [T]} \max(0, p_{t} - p_{\sigma(t)}^{\ins}), \quad \mathcal{L}_{\incor\_\ins} = \frac{1}{T}\sum_{ t\in [T]} \max(0, p_{\sigma(t)}^{\ins} - p_{t})
    \mathcal{L}_{\regu\shyp\cor\_\ins} &= \mathbb{E}_{t\in [T]}[\max(0, p_{t} - p_{\sigma(t)}^{\ins})], \\ \mathcal{L}_{\regu\shyp\incor\_\ins} &= \mathbb{E}_{t\in [T]}[\max(0, p_{\sigma(t)}^{\ins} -p_{t})],
\end{align}
where $\mathcal{L}_{\regu\shyp\cor\_\ins}$ and $\mathcal{L}_{\regu\shyp\incor\_\ins}$ are losses for augmented interaction sequences of inserting correctly and incorrectly answered interactions, respectively.

\vspace{-0.05in}
\subsection{Deletion}

Similar to the insertion augmentation strategy, we enforce monotonicity by removing some interactions in the original interaction sequence based on the following hypothesis: if a student's response record contains less correct answers, the correctness probabilities for the remaining questions would become decrease and vice versa.
Formally, from the original interaction sequence $(I_1, \dots, I_T)$, we randomly sample a set of indices $\mathbf{D} \subset [T]$, where $R_{t} = 1$ (resp. $R_{t} = 0$) for $t \in \mathbf{D}$, based on the Bernoulli distribution with the probability $\alpha_{\del}$.
We remove the index $t \in \mathbf{D}$ and impose the hypothesis $p_{t} \geq p_{\sigma(t)}^{\del}$ (resp. $p_{t} \leq p_{\sigma(t)}^{\del}$), where $p_t$ and $p_{t}^{\del}$ are model's predicted correctness probabilities for $t$-th question of the original and augmented sequences, respectively.  
Here, $\sigma: [T] - \mathbf{D} \to [T']$ is the order preserving bijection with $I_t = I_{\sigma(t)}^{\del}$ for $t \in [T] -\mathbf{D}$.
We impose the hypothesis through the following losses:
\begin{align}
    % \mathcal{L}_{\cor\_\del} = \frac{1}{T'} \sum_{t\in [T] -\mathbf{D}} \max(0, p_{\sigma(t)}^{\del} - p_{t}), \quad \mathcal{L}_{\incor\_\del} = \frac{1}{T'} \sum_{t\in [T]-\mathbf{D}} \max(0, p_t - p_{\sigma(t)}^{\del})
    \mathcal{L}_{\regu\shyp\cor\_\del} = \mathbb{E}_{t\not\in \mathbf{D}}[\max(0, p_{\sigma(t)}^{\del} - p_{t})], \\ \mathcal{L}_{\regu\shyp\incor\_\del} = \mathbb{E}_{t\not\in \mathbf{D}}[\max(0, p_t - p_{\sigma(t)}^{\del})], 
\end{align}
where $\mathcal{L}_{\regu\shyp\cor\_\del}$ and $\mathcal{L}_{\regu\shyp\incor\_\del}$ are losses for augmented interaction sequences of deleting correctly and incorrectly answered interactions, respectively.

% \subsection{Total loss}

% INS+DEL, INS+DEL+REP VERSION

\begin{table}[t]
\vspace{-0.1in}
\centering
\resizebox{\columnwidth}{!}
{
\small
\begin{tabular}{cccccc}\toprule
dataset & model & no augmentation &  
\adjustbox{stack=ll}{insertion \\ + deletion} &
% insertion + deletion &
\adjustbox{stack=ll}{insertion\\+ deletion\\+ replacement}\\ 
% insertion + deletion + replacement \\
\midrule
\multirow{3}{*}{ASSIST2015} & DKT & \auc{72.01}{0.05} &  \boldaucgain{72.46}{0.06}{0.6} & \aucgain{72.39}{0.07}{0.5} \\
& DKVMN & \auc{71.21}{0.09} & \aucgain{72.00}{0.18}{1.1} & \boldaucgain{72.23}{0.09}{1.4} \\
% & SAKT & \\}
& SAINT & \auc{72.13}{0.09} & \aucgain{72.78}{0.06}{0.9}  & \boldaucgain{72.81}{0.04}{0.9} \\ \midrule
\multirow{3}{*}{ASSISTChall} & DKT & \auc{74.40}{0.16} & \aucgain{75.98}{0.07}{2.1} & \boldaucgain{79.07}{0.08}{6.3} \\
& DKVMN & 74.46 $\pm$ \tiny{0.11} & \aucgain{75.06}{0.10}{0.8} & \boldauc{78.21}{0.05} \\
% & SAKT & \\
& SAINT & \auc{77.01}{0.18} & \aucgain{78.02}{0.09}{1.0} &\boldauc{80.18}{0.05} \\ \midrule
\multirow{3}{*}{STATICS2011} & DKT & \auc{86.43}{0.29} & \aucgain{87.18}{0.12}{0.7}  & \boldaucgain{87.27}{0.11}{1.0} \\
& DKVMN & 84.89 $\pm$ \tiny{0.17} & \aucgain{85.65}{0.94}{0.9} & \boldaucgain{87.17}{0.14}{2.7} \\
% & SAKT & \\
& SAINT & \auc{85.82}{0.50} & \aucgain{86.53}{0.30}{1.0} & \boldaucgain{87.56}{0.06}{2.2}\\ \midrule
% JunYi & & & & \\
% KDDCup2010 & & &  & & & & \\
\multirow{3}{*}{EdNet-KT1} & DKT & \auc{72.75}{0.09} & \aucgain{74.04}{0.04}{1.8} & \boldaucgain{74.28}{0.06}{2.1} \\
& DKVMN & \auc{73.58}{0.08} & \aucgain{73.94}{0.05}{0.5} & \boldaucgain{74.16}{0.11}{0.8} \\
% & SAKT & \\
& SAINT & \auc{74.78}{0.05} & \boldauc{75.32}{0.05} & \auc{75.26}{0.02}  \\
\bottomrule
\end{tabular}
}
\caption{Test AUCs of DKT, DKVMN, and SAINT models on 4 public benchmark datasets. 
The results show the mean and standard deviation averaged over 5 runs and the best result for each dataset and model is indicated in bold. }
\vspace{-0.15in}
\label{tab:main}
\end{table}

\vspace{-0.05in}
\section{Experiments} \label{sec:exp}

%\vspace{-0.05in}
%\subsection{Experimental setup}

We demonstrated the effectiveness of the proposed method on 4 widely used benchmark datasets: ASSISTments2015 \cite{feng2009addressing}, ASSISTmentsChall, STATICS2011, and EdNet-KT1 \cite{choi2020ednet}. 
We describe the details about these benchmarks, their statistics, and pre-processing procedures in the Appendix.
% {ASSISTments} datasets are the most widely used benchmark for Knowledge Tracing, which is provided by ASSISTments online tutoring platform\footnote{https://new.assistments.org/} \cite{feng2009addressing}. There are several versions of dataset depend on the years they collected, and we used ASSISTments2015\footnote{https://sites.google.com/site/assistmentsdata/home/2015-assistments-skill-builder-data} and ASSISTmentsChall\footnote{https://sites.google.com/view/assistmentsdatamining}. 
% ASSISTmentsChall dataset is provided by the 2017 ASSISTments data mining competition. 
% {STATICS2011} consists of the interaction logs from an engineering statics course, which is available on the PSLC datashop\footnote{https://pslcdatashop.web.cmu.edu/DatasetInfo?datasetId=507}. 
% {EdNet-KT1} is the largest publicly available interaction dataset consists of TOEIC (Test of English for Interational Communication) problem solving logs collected by \emph{Santa}\footnote{https://aitutorsanta.com/} \cite{choi2020ednet}.
% We reduce the size of the EdNet-KT1 dataset by sampling 6000 users among 600K users.
% Detailed statistics and pre-processing methods for these datasets are described in Appendix. 
% With the exception of the EdNet-KT1 dataset, we used 80\% of the students as a training set and the remaining 20\% as a test set. 

%\textbf{Details for training and evaluation.}
We test DKT \cite{piech2015deep}, DKVMN \cite{zhang2017dynamic}, and SAINT \cite{choi2020towards} models. % for experiments. 
For DKT, we set the embedding dimension and the hidden dimension as 256. 
For DKVMN, key, value, and summary dimension are all set to be 256, and we set the number of latent concepts as 64.
% SAKT has 1 layer with hidden dimension 256. 
SAINT has 2 layers with hidden dimension 256, 8 heads, and feed-forward dimension 1024. 
All the models do not use any additional features of interactions except question ids and responses as an input, and the model weights are initialized with Xavier distribution \cite{glorot2010understanding}. 
They are trained from scratch with batch size 64, and we use the Adam optimizer 
% \cite{kingma2014adam} 
with learning rate $0.001$ which is scheduled by Noam scheme with warm-up step 4000. 
We set each model's maximum sequence size as 100 on ASSISTments2015 \& EdNet-KT1 dataset and 200 on ASSISTmentsChall \& STATICS2011 dataset. 
% We also use Noam scheduling with warmup step set to 4000   \cite{vaswani2017attention}. 
Hyperparameters for augmentations, $\alpha_{\aug}, \lambda_{\regu\shyp\aug}$, and $\lambda_{\aug}$  are searched over $\alpha_{\aug} \in \{0.1, 0.3, 0.5\}$, $\lambda_{\regu\shyp\aug} \in \{1, 10, 50, 100\}$, and $\lambda_{\aug} \in \{0, 1\}$. 
% We use $\lambda_{\aug} = 1$ for the most of the experiments. 
For all dataset, we evaluate our results using 5-fold cross validation and use Area Under Curve (AUC) as an evaluation metric.
% , which is the most commonly used evaluation metric for KT. 

% \subsubsection{Evaluation}

% 100 step val set evaluation -> model with max auc

% DKT, ASSISTCHALL
% \begin{table}[!htp]\centering
% \resizebox{\columnwidth}{!}{
% \scriptsize
% \label{tab:consablation}
% \begin{tabular}{lccccc}\toprule
% & replacement & correct insertion & incorrect insertion & correct deletion & incorrect deletion \\ \midrule
% train with the loss (\ref{eqn:onlyaug}) & \auc{75.13}{0.04}  &  \auc{74.61}{0.17} & \auc{74.57}{0.14} & \auc{74.92}{0.12} & \auc{74.42}{0.20} \\
% % constraint & 78.85 $\pm$ 0.09 & 76.98 & 77.62 & 77.50 & 77.24 \\
% % both & 79.05 & 77.61 & 77.93 & 78.04 & 76.07 \\ \hdashline[0.5pt/2pt]
% train with the loss (\ref{eqn:total_loss}) & \auc{78.85}{0.08} & \auc{75.98}{0.07} & 75.64 $\pm$ \tiny{0.12} & \auc{75.58}{0.06} & \auc{74.71}{0.16} \\
% % \hdashline[0.5pt/2pt]
% % no\_augmentation & \multicolumn{5}{c}{74.40 $\pm$ 0.16} \\
% \bottomrule
% \end{tabular}
% }
% \caption{Comparison of the performances (AUC) of the DKT model on ASSISTChall dataset, trained with only data augmentation (\eqref{eqn:onlyaug}) and with consistency and monotonicity regularizations (\eqref{eqn:total_loss}). Note that the AUC without augmentation is 74.40. \textcolor{orange}{Better row name?}}
% % Training the model only with data augmentation is not enough for the meaningful improvement in AUC, but including consistency and monotonicity regularization losses boost up the performances.  
% \label{tab:cons}
% \end{table}

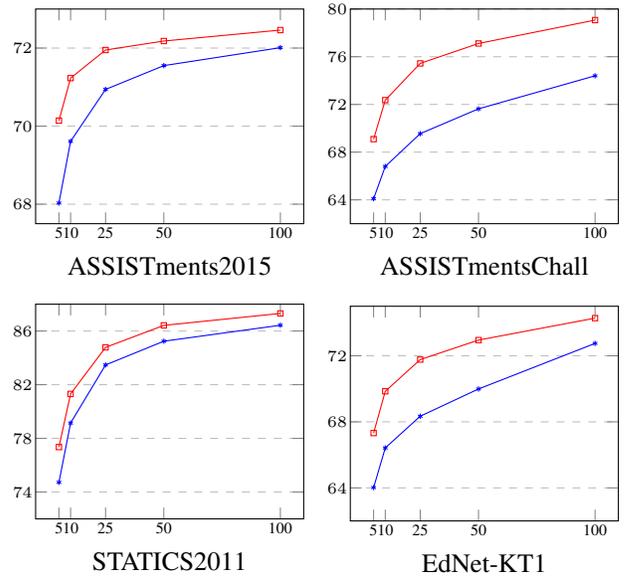
\begin{figure}[t]
\vspace{-0.15in}
\centering
\begin{tikzpicture}[every mark/.append style={mark size=1pt}]
\begin{axis}[
    title style={at={(0.5,0)},anchor=north,yshift=-15.0},
    title={ASSISTments2015},
    % width=0.30\linewidth,
    width=0.6\columnwidth,
    at={(-220,0)},
    xmin=-5, xmax=110,
    ymin=67.5, ymax=73,
    xtick={5, 10, 25, 50, 100},
    xticklabels={5, 10, 25, 50, 100},
    ytick={66, 68, 70, 72, 74},
    legend pos=south east,
    ymajorgrids=true,
    grid style=dashed,
    legend style={nodes={scale=0.6}},
    yticklabel style = {font=\tiny,xshift=0.5ex},
    xticklabel style = {font=\tiny,yshift=0.5ex}
    % legend image post style={mark=*}
]
\addplot[
    color=red,
    mark=square,
    ]
    coordinates {
    (5,70.14)(10,71.23)(25,71.95)(50,72.18)(100,72.46)
    };
\addplot[
    color=blue,
    mark=asterisk,
    ]
    coordinates {
    (5,68.03)(10,69.61)(25,70.94)(50,71.55)(100,72.01)
    };
% \legend{w/ augmentation, w/o augmentation}
% \legend{no augmentation, insertion, deletion}
\end{axis}

% assistchall
\begin{axis}[
    title style={at={(0.5,0)},anchor=north,yshift=-15.0},
    title={ASSISTmentsChall},
    % width=0.30\linewidth,
    width=0.6\columnwidth,
    at={(-85,0)},
    xmin=-5, xmax=110,
    ymin=62, ymax=80,
    xtick={5, 10, 25, 50, 100},
    xticklabels={5, 10, 25, 50, 100},
    ytick={60,64,68,72,76,80},
    legend pos=south east,
    ymajorgrids=true,
    grid style=dashed,
    legend style={nodes={scale=0.6}},
    yticklabel style = {font=\tiny,xshift=0.5ex},
    xticklabel style = {font=\tiny,yshift=0.5ex}
    % legend image post style={mark=*}
]
% rep + ins + del
\addplot[
    color=red,
    mark=square,
    ]
    coordinates {
    (5,69.09)(10,72.36)(25,75.44)(50,77.11)(100,79.07)
    };
\addplot[
    color=blue,
    mark=asterisk,
    ]
    coordinates {
    (5,64.1)(10,66.80)(25,69.55)(50,71.63)(100,74.40)
    };

% \legend{no augmentation, replacement, insertion, deletion}
% \legend{w/ augmentation, w/o augmentation}
\end{axis}

% statics
\begin{axis}[
    % width=0.30\linewidth,
    width=0.6\columnwidth,
    at={(-220, -220)},
    title style={at={(0.5,0)},anchor=north,yshift=-15.0},
    title={STATICS2011},
    xmin=-5, xmax=110,
    ymin=72, ymax=88,
    xtick={5, 10, 25, 50, 100},
    xticklabels={5, 10, 25, 50, 100},
    ytick={74, 78, 82, 86, 90},
    legend pos=south east,
    ymajorgrids=true,
    grid style=dashed,
    legend style={nodes={scale=0.6}},
    yticklabel style = {font=\tiny,xshift=0.5ex},
    xticklabel style = {font=\tiny,yshift=0.5ex}
    % legend image post style={mark=*}
]
% rep + ins + del
\addplot[
    color=red,
    mark=square,
    ]
    coordinates {
    % (5,76.23)(10,79.75)(25,83.93)(50,85.67)(100,86.97)
    (5,77.35)(10,81.31)(25,84.78)(50,86.42)(100,87.31)
    };
\addplot[
    color=blue,
    mark=asterisk,
    ]
    coordinates {
    (5,74.72)(10,79.14)(25,83.47)(50,85.24)(100,86.43)
    };

% \legend{no augmentation, replacement, insertion, deletion}
% \legend{w/ augmentation, w/o augmentation}
\end{axis}

% ednet

\begin{axis}[
    title style={at={(0.5,0)},anchor=north,yshift=-15.0},
    % width=0.30\linewidth,
    width=0.6\columnwidth,
    at={(-85,-180)},
    title={EdNet-KT1},
    xmin=-5, xmax=110,
    ymin=62, ymax=75,
    xtick={5, 10, 25, 50, 100},
    xticklabels={5, 10, 25, 50, 100},
    ytick={64, 68, 72, 76},
    legend pos=south east,
    ymajorgrids=true,
    grid style=dashed,
    legend style={nodes={scale=0.5}},
    yticklabel style = {font=\tiny,xshift=0.5ex},
    xticklabel style = {font=\tiny,yshift=0.5ex}
    % legend image post style={mark=*}
]
% rep + ins + del
\addplot[
    color=red,
    mark=square,
    ]
    coordinates {
    (5,67.32)(10, 69.85)(25,71.77)(50,72.95)(100, 74.28)
    };
\addplot[
    color=blue,
    mark=asterisk,
    ]
    coordinates {
    (5,64.02)(10,66.42)(25,68.34)(50,69.99)(100,72.75)
    };
% \legend{w/ augmentation, w/o augmentation}
\end{axis}

\end{tikzpicture}
\caption{Test AUCs with various sizes of training data under the DKT model. $x$ axis stands for the portion of the training set we use for training (relative to the full train set) and $y$ axis is the AUC. 
Blue line represents the AUCs of the vanilla DKT model, and red line represents the AUCs of the DKT model trained with augmentations and regularizations. }
\label{fig:smalldata}
\vspace{-0.1in}
\end{figure}

%\subsection{Results}

\vspace{-0.05in}
\subsection{Main results}

\begin{figure}
\vspace{-0.05in}
    \centering
        \includegraphics[width=3.3in]{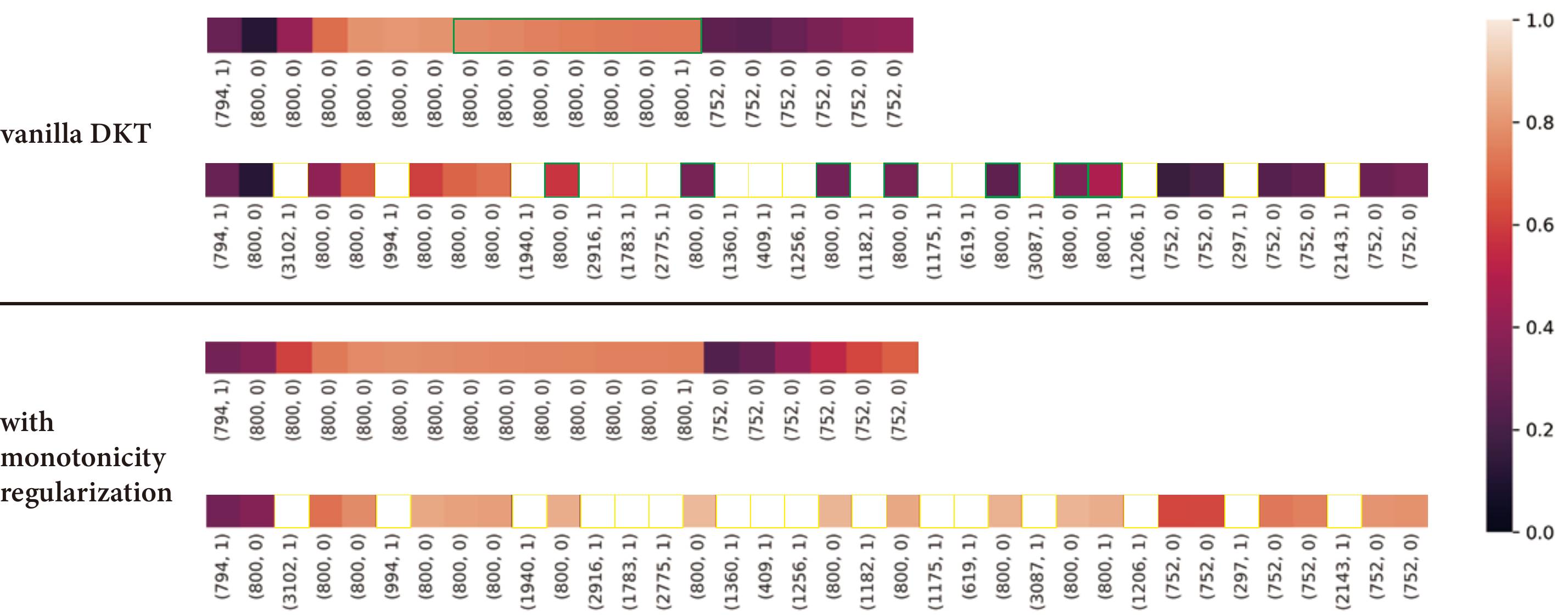}
    \caption{
    Response correctness prediction for a student in the ASSISTmentsChall dataset. We randomly insert interactions with correct responses (interactions with yellow boundaries).
    In case of the vanilla DKT model, the predictions for the original interactions (especially the interactions with green boundaries) are decreased, even if the student answered more questions correctly. 
    However, such problem is resolved when we train the model with monotonicity regularization.
    % (with the loss $\mathcal{L}_{\tot} = \mathcal{L}_{\ori} + \mathcal{L}_{\cor\_\ins} + 100 \cdot\mathcal{L}_{\regu\shyp\cor\_\ins}$). 
    Unlike the vanilla DKT model, predicted correctness probabilities for the original interactions are increased after insertion.
    }
    \label{fig:heatmap}
    \vspace{-0.15in}
\end{figure}

The results (AUCs) are shown in Table \ref{tab:main} that compares models without and with augmentations, and we report the best results for each strategy. 
(The detailed hyperparameters for these results are given in Supplementary materials.)
The 4th column represents results using both insertion and deletion, and the last column shows the results with all 3 augmentations. 
Since there's no big difference on performance gain between insertion and deletion, we only report the performance that uses one or both of them together.
We use skill-based replacement if skill information for each question in the dataset is available, and use question-random replacement that that selects new questions among all questios if not (e.g. ASSISTments2015). 
As one can see, the models trained with consistency and monotonicity regularizations outperforms the models without augmentations in a large margin, regardless of model's architectures or datasets.
Using all three augmentations gives the best performances for most of the cases. 
{For instance, there exists 6.3\% gain in AUC on ASSISTmentsChall dataset under the DKT model. }
{Furthermore, not only enhancing the prediction performances, our training scheme also resolves the vanilla model's issue where the monotonicity condition on the predictions of original and augmented sequences is violated. 
As in Figure \ref{fig:heatmap}, the predictions of the model trained with monotonicity regularization (correct insertion) are increased after insertion, which contrasts to the vanilla DKT model's outputs.}

%\textbf{Training with smaller dataset.}
% \textbf{Training with smaller dataset.}
% Overfitting tends to be more serious when we train a model on smaller datasets. 
Since overfitting is expected to be more severe when using a smaller dataset, we conduct experiments using various fractions of the existing training datasets (5\%, 10\%, 25\%, 50\%) and show that our augmentations yield more significant improvements for smaller training datasets. 
Figure \ref{fig:smalldata} shows performances of DKT model on various datasets, with and without augmentations.
For example, on ASSISTmentsChall dataset, using 100\% of the training data gives AUC 74.4\%, while the same model trained with augmentations achieved AUC 75.44\% with only 25\% of the training dataset.

% ALL DATASET
\begin{table*}[t]
% \vspace{-0.1in}
\centering
% \resizebox{\columnwidth}{!}
{
\small
\label{tab:consablation}
\begin{tabular}{ccccccc}\toprule
% dataset &  loss & replacement & \adjustbox{stack=cc}{correct\\ insertion} & \adjustbox{stack=cc}{incorrect\\ insertion} & \adjustbox{stack=cc}{correct\\ deletion} & \adjustbox{stack=cc}{incorrect\\ deletion} \\  \midrule
dataset &  loss & replacement & insertion, O & insertion, X & deletion, O & deletion, X \\  \midrule
ASSIST2015 & \eqref{eqn:onlyaug}& \auc{72.03}{0.06} & \auc{71.98}{0.06} & \auc{71.98}{0.05} & \auc{72.05}{0.04} & \auc{72.04}{0.02} \\
\cmidrule{2-7} 
(\auc{72.01}{0.05}) & \eqref{eqn:total_loss}
 & \auc{72.06}{0.03} & \auc{72.09}{0.06} & \auc{72.35}{0.11} & \auc{72.53}{0.08} & \auc{72.26}{0.04}  \\
\midrule
ASSISTChall & \eqref{eqn:onlyaug}&  \auc{75.13}{0.04}  &  \auc{74.61}{0.17} & \auc{74.57}{0.14} & \auc{74.92}{0.12} & \auc{74.42}{0.20} \\
\cmidrule{2-7}  
(\auc{74.40}{0.16}) & \eqref{eqn:total_loss}
&  \auc{78.85}{0.08} & \auc{75.98}{0.07} & \auc{75.64}{0.12} & \auc{75.60}{0.06} & \auc{74.77}{0.11} \\
\midrule
STATICS2011 & \eqref{eqn:onlyaug}& \auc{86.89}{0.23} & \auc{86.45}{0.26} & \auc{86.40}{0.22} & \auc{86.53}{0.29} & \auc{86.55}{0.25} \\
\cmidrule{2-7} 
(\auc{86.43}{0.29}) & \eqref{eqn:total_loss} 
& \auc{87.27}{0.11} & \auc{86.72}{0.23} & \auc{87.18}{0.12} & \auc{87.07}{0.33} & \auc{86.97}{0.26} \\
\midrule
EdNet-KT1 & \eqref{eqn:onlyaug}& \auc{73.04}{0.10} & \auc{72.81}{0.08} & \auc{72.88}{0.09} & \auc{72.99}{0.07} & \auc{73.28}{0.04} \\
\cmidrule{2-7} 
(\auc{72.75}{0.09})& \eqref{eqn:total_loss} 
& \auc{73.89}{0.06} & \auc{73.73}{0.06} & \auc{73.52}{0.06} & \auc{74.04}{0.04} & \auc{73.76}{0.04} \\
\bottomrule
\end{tabular}
}
\caption{Comparison of the test AUCs of the DKT model, trained with only data augmentation (i.e., using the loss \eqref{eqn:onlyaug}) and with consistency and monotonicity regularizations (i.e., using the loss \eqref{eqn:total_loss}). 
AUCs of the vanilla DKT model are given in parentheses below the dataset names. O (resp. X) represents correct (resp. incorrect) response.}
\label{tab:cons}
\vspace{-0.1in}
\end{table*}

\vspace{-0.05in}
\subsection{Ablation study}

%

% \subsubsection{Are constraint losses necessary?}
\textbf{Are constraint losses necessary?} 
One might think that data augmentations are enough for boosting up the performance, and imposing consistency and monotonicity are not necessary. 
However, we found that including such regularization losses for training is essential for further performance gain. 
To see this, we compare the performances of the model trained only with KT losses for both original and augmented sequences 
\begin{align}
\label{eqn:onlyaug}
    \mathcal{L}_{\tot} = \mathcal{L}_{\ori} + \sum_{\aug \in \mathcal{A}}\lambda_{\aug}\mathcal{L}_{\aug}
\end{align}
(where $\lambda_{\aug} = 1$) and with consistency and monotonicity regularization losses \eqref{eqn:total_loss}
% \begin{align}
% \label{eqn:both}
%     \mathcal{L}_{\tot} = \mathcal{L}_{\ori} + \sum_{\aug\in \mathcal{A}}(\lambda_{\aug} \mathcal{L}_{\aug} + \lambda_{\regu\shyp\aug}\mathcal{L}_{\regu\shyp\aug})
% \end{align}
where $\mathcal{A}$ is a set that contains a single augmentation.
Training a model with the loss \eqref{eqn:onlyaug} can be thought as using augmentations without imposing any consistency or monotonicity biases. 
% For replacement and insertion, we do not include new interactions in $\mathcal{L}_{\KT}^{\aug}$. 
% For each augmentations, we compared results with three different total losses: 
% \begin{enumerate}
%     \item $\mathcal{L}_{\tot} = \mathcal{L}_{\KT} + \mathcal{L}_{\KT}^{\aug}$ (include KT loss of augmented sequences)
%     \item $\mathcal{L}_{\tot} = \mathcal{L}_{\KT} + \lambda_{\aug}\mathcal{L}_{\aug}$ (include constraint loss)
%     \item $\mathcal{L}_{\tot} = \mathcal{L}_{\KT} + \lambda_{\aug} \mathcal{L}_{\aug} + \mathcal{L}_{\KT}^{\aug}$ (include both KT loss and constraint loss) 
% \end{enumerate}

% \textbf{Other deep KT models}

Table \ref{tab:cons} shows results under the DKT model.
{Using only data augmentation (training the model with the loss (\ref{eqn:onlyaug})) gives a marginal gain in performances or even worse performances. 
However, training with both data augmentation and consistency or monotonicity regularization losses (\ref{eqn:total_loss}) give significantly higher performance gain. 
Under ASSISTmentsChall dataset, using replacement along with consistency regularization improves AUC by 6\%, which is much higher than the 1\% improvement only using data augmentation. }

% \subsubsection{Ablation on monotonicity constraints}
\textbf{Ablation on monotonicity constraints} 
We perform an ablation study to compare the effects of monotonicity regularization and \emph{reversed} monotonocity regularization. Monotonocity regularization introduces constraint loss to align the inserted or deleted sequence in order to modify the probability of correctness of the original sequence to follow insertion or deletion. For example, when a correct response is inserted to the sequence, the probability of correctness for the original sequence increases. Reversed monotonocity regularization modifies the probability of correctness in the opposite manner, where inserting a correct response would decrease the probability of correctness in the original sequence.

For each $\aug\in \{\cor\_\ins, \incor\_\ins, \cor\_\del, \incor\_\del\}$, we can define reversed version of the monotonicity regularization loss 
$\mathcal{L}^{\mathrm{rev}}_{\regu\shyp\aug}$  which impose the opposite constraint on the model's output, e.g. we define $\mathcal{L}_{\regu\shyp\cor\_\ins}^{\mathrm{rev}}$ as

% \begin{equation}
% \resizebox{.91\linewidth}{!}{$
%     \displaystyle
%     x = \prod_{i=1}^n \sum_{j=1}^n j_i + \prod_{i=1}^n \sum_{j=1}^n i_j + \prod_{i=1}^n \sum_{j=1}^n j_i + \prod_{i=1}^n \sum_{j=1}^n i_j + \prod_{i=1}^n \sum_{j=1}^n j_i
% $}
% \end{equation}%
\vspace{-0.15in}
\begin{align}
\resizebox{.91\linewidth}{!}{
$\mathcal{L}_{\regu\shyp\cor\_\ins}^{\mathrm{rev}} = \mathbb{E}_{t\in [T]}[\max(0, p_{\sigma(t)}^{\ins} - p_{t})] =\mathcal{L}_{\regu\shyp\incor\_\ins}$ 
}
\end{align}
which forces model's output of correctness probability to \emph{decrease} when \emph{correct} responses are inserted. 
In this experiments, we do not include KT loss for augmented sequences (set $\lambda_{\aug} =0$) to observe the effects of consistency loss only.
Also, the same hyperparameters ($\alpha_{\aug}$ and $\lambda_{\regu\shyp\aug}$) are used for both the original and reversed constraints.

\begin{table}[t]\centering
\resizebox{\columnwidth}{!}
% \vspace{-0.15in}
{
\small
\label{tab:consablation}
\begin{tabular}{cccccc}\toprule
augmentation & direction & ASSIST2015 & ASSISTChall & STATICS2011 & EdNet-KT1  \\ \midrule
- & - & \auc{72.01}{0.05}  & \auc{74.40}{0.16} & \auc{86.43}{0.29} & \auc{72.75}{0.09}\\ 
% \hdashline[0.5pt/2pt]
\midrule
insertion, O & \upar & \auc{72.08}{0.02} & \auc{75.98}{0.06} & \auc{86.72}{0.23} & \auc{73.70}{0.08} \\
insertion, X  & \doar & \auc{72.31}{0.04} & \auc{75.34}{0.16} & \auc{87.18}{0.12} & \auc{73.40}{0.06} \\ \hdashline[0.5pt/2pt]
deletion, O & \doar & \auc{72.44}{0.05} & \auc{75.60}{0.06} & \auc{87.07}{0.33} & \auc{74.01}{0.05} \\
deletion, X & \upar & \auc{72.26}{0.04} & \auc{74.77}{0.11} & \auc{86.68}{0.27} & \auc{73.71}{0.04}\\ 
 %\hdashline[0.5pt/2pt]
 \midrule
insertion, O & \doar & \auc{71.79}{0.06} & \auc{75.42}{0.17} & \auc{86.58}{0.50} & \auc{69.67}{0.06} \\
insertion, X & \upar & \auc{70.73}{0.10} & \auc{74.92}{0.11} & \auc{86.22}{0.18} & \auc{71.95}{0.15} \\ \hdashline[0.5pt/2pt]
deletion, O & \upar & \auc{66.48}{0.10} & \auc{74.68}{0.13} & \auc{86.76}{0.27} & \auc{71.23}{0.81} \\
deletion, X & \doar & \auc{67.34}{0.17} & \auc{73.91}{0.14} & \auc{86.58}{0.28} & \auc{69.99}{0.11} \\
\bottomrule
\end{tabular}
}
\caption{Ablation test on the directions of monotonicity regularizations with the DKT model. 2nd to 5th rows show the results with the original regularization losses, and the last 4 rows show the results with the reversed regularization losses. \upar\, (resp. \doar) means that the loss impose increasing (resp. decreasing) bias on a model.}
\label{tab:monotonicity}
\vspace{-0.1in}
\end{table}

% SMALL TABLE
\begin{table}[t]
% \vspace{-0.05in}
\centering
\label{tab:repablation}
\resizebox{\columnwidth}{!}
{
\small
\begin{tabular}{cccccc}
\toprule
dataset &  vanilla &  replaced inters & remaining inters & full inters \\ \midrule
ASSIST2015 & \auc{72.01}{0.05} & \auc{70.53}{0.07} & \boldauc{72.07}{0.02} & \auc{71.39}{0.09}  \\ 
ASSISTChall  & \auc{74.40}{0.16} & \auc{74.68}{0.09} & \boldauc{78.45}{0.08} & \auc{75.91}{0.07}  \\ 
STATICS2011  & \auc{86.43}{0.29} & \auc{82.97}{0.27} & \boldauc{87.17}{0.15} & \auc{83.49}{0.10} \\
EdNet-KT1 & \auc{72.75}{0.09} & \auc{65.52}{0.07} & \boldauc{73.87}{0.10} & \auc{68.77}{0.11}  \\
\bottomrule
\end{tabular}
}
\caption{{Test AUCs of the DKT model with variations of replacements and qDKT with Lapacian regularization}. Best result for each dataset is indicated in bold.}
% For ASSISTments2015 dataset, the results with question-random replacement are reported. Best result for each dataset is indicated in bold.}
\label{tab:rep_location}
\vspace{-0.1in}
\end{table}

% SMALL TABLE - DKT ONLY
\begin{table*}[t]\centering
\label{tab:repablation}
% \resizebox{\columnwidth}{!}
{
\small
\begin{tabular}{cccccc}\toprule
dataset & no augmentation & question-random & interaction-random & skill-set & skill \\ \midrule
ASSIST2015 & \auc{72.01}{0.05} & \boldauc{72.07}{0.02} & \auc{71.77}{0.05} & - & -  \\
ASSISTChall  & \auc{74.40}{0.16} & \auc{78.39}{0.09} & \auc{74.27}{0.38} & \auc{77.57}{0.08}& \boldauc{78.45}{0.08}   \\ 
STATICS2011 & \auc{86.43}{0.29} & \auc{86.35}{0.06} & \auc{84.50}{0.28} & - & \boldauc{87.17}{0.15} \\ 
EdNet-KT1 & \auc{72.75}{0.09} & \auc{73.84}{0.05} & \auc{72.62}{0.17} & \auc{73.80}{0.07}& \boldauc{73.87}{0.10} \\
\bottomrule
\end{tabular}
}
\caption{Test AUCs of the DKT model with different type of replacements - question-random replacement, interaction-random replacement, skill-set-based replacement, and skill-based replacement. 
% Note that skill information is not available for ASSIST2015 dataset. 
Best result for each dataset is indicated in bold.}
\label{tab:rep_skill}
\vspace{-0.1in}
\end{table*}

% SMALL TABLE
\begin{table*}[t]
% \vspace{-0.05in}
\centering
% \resizebox{0.7\columnwidth}{!}
{
\small
\begin{tabular}{ccccc}
\toprule
dataset & DKT & DKT+ & qDKT & ours \\ \midrule
ASSIST2015 & \auc{72.01}{0.05} & \auc{72.28}{0.08} & - & \boldauc{72.46}{0.06} \\ 
ASSISTChall & \auc{74.40}{0.16} & \auc{74.70}{0.08} & \auc{75.03}{0.08} & \boldauc{79.07}{0.08}  \\ 
STATICS2011 & \auc{86.43}{0.29} & \auc{85.66}{0.08} & \auc{86.65}{0.12} & \boldauc{87.27}{0.11} \\
EdNet-KT1 & \auc{72.75}{0.09} & \auc{73.43}{0.05} & \auc{69.77}{0.22} & \boldauc{74.28}{0.06}\\
\bottomrule
\end{tabular}
}
\caption{Test AUCs of the vanilla DKT, DKT+, qDKT, and DKT with our regularizations. Best result for each dataset is indicated in bold. Detailed hyperparameters are given in the Appendix.}
% For ASSISTments2015 dataset, the results with question-random replacement are reported. Best result for each dataset is indicated in bold.}
\label{tab:comparison}
\vspace{-0.15in}
\end{table*}

Table \ref{tab:monotonicity} shows the performances of DKT model 
% on \textcolor{blue}{ASSIST2015 dataset} 
with the original and reversed monotonicity regularizations.
Second row represents the performance with no augmentations, the 3rd to the 6th rows represent the results from using original (aligned) insertion/deletion monotonicity regularization losses, and the last four rows represent the results when the reversed monotonicity regularization losses are used. 
The results demonstrate that using aligned monotonicity regularization loss outperforms the model with reversed monotonicity regularization loss. 
Also, the performances of reversed monotonicity shows large decrease in performance on several datasets even compared to the model with no augmentation. 
{In case of the EdNet-KT1 dataset, the model's performance with correct insertion along with original regularization improves the AUC from 72.75\% to 73.70\%, while using the reversed regularization drops the performance to 69.67\%. }
% \textcolor{blue}{This shows that our monotonicity constraints are working as meaningful restrictions.} 

% \subsubsection{Ablation on replacement}
\textbf{Ablation on replacement}
We compare our consistency regularization with the other two variations of replacements, consistency regularization on replaced interactions and full interactions, correspond to the following losses:
% - which correspond to the following losses respectively:
\vspace{-0.05in}
% We compare our consistency regularization with the other two variations of replacements - consistency regularization on replaced interactions and full interactions -  and qDKT  \cite{sonkar2020qdkt}.
% As we mentioned in Section 3, there are two more possible variations of the consistency loss for the replacement depends on whether we include replaced interaction's output in the loss or not:
\begin{align}
\mathcal{L}_{\regu\shyp\rep\_\mathrm{ro}} &= \mathbb{E}_{t\in \mathbf{R}}[(p_{t} - p_{t}^{\rep})^{2}],\\ \mathcal{L}_{\regu\shyp\rep\_\mathrm{full}} &= \mathbb{E}_{t\in [T]}[(p_{t} - p_{t}^{\rep})^{2}],
\label{eqn:rep_var}
\end{align}
where $\mathrm{ro}$ stands for \emph{replaced only}. 
We compared such variations with the original consistency loss $\mathcal{L}_{\regu\shyp\rep}$ that does not include predictions for the replaced interactions. 
For all variations, we used the same replacement probability $\alpha_{\rep}$ and loss weight $\lambda_{\regu\shyp\rep}$, and we do not include KT loss for replaced sequences  as before.
% Also, we compare replacement with qDKT that uses the following Laplacian loss which regularizes the variance of predicted correctness probabilities for questions that fall under the same skill: 
% \begin{align}
%     \mathcal{L}_{\mathrm{Laplacian}} = \mathbb{E}_{(q_i, q_j) \in \mathcal{Q}\times\mathcal{Q}} [\bm{1}(i, j)(p_{i} - p_{j})^{2}]
% \end{align}
% where $\mathcal{Q}$ is the set of all questions, $p_i, p_j$ are the model's predicted correctness probabilities for the questions $q_i, q_j$, and $\bm{1}(i, j)$ is 1 if $q_i, q_j$ have common skills attached, otherwise 0. 
% {It is similar to our variation of consistency regularization ($\mathcal{L}_{\regu\shyp\rep\_\mathrm{ro}}$ in \eqref{eqn:rep_var}) that only compares replaced interactions' outputs, but it does not replace questions and it compares all questions (with same skills) at once.} 
% Since the hyperparameter $\lambda$ that scales the Laplacian regularization term is not provided in the paper, we use the same set of hyperparameters we use for other losses, and report the best results among them. 
Table \ref{tab:rep_location} shows that including the replaced interactions' outputs hurt performances. 
% For example, under the EdNet-KT1 dataset, all the variations of consistency regularization and Laplacian regularization significantly dropped AUCs to under 70\%, while the original consistency regularization boost up the performance from 72.75\% to 73.87\%.  

% To see the effect of using the skill information of questions for replacement, we compared skill-based replacement with two different random versions of replacement: \emph{question random replacement} and \emph{interaction random replacement}. 
To see the effect of using the skill information of questions for replacement, we compared skill-based replacement with three different random versions of replacement: \emph{question random replacement}, \emph{interaction random replacement}, and \emph{skill-set-based replacement}. 
For \emph{question random replacement}, we replace questions with different ones randomly (without considering skill information), while \emph{interaction random replacement} changes both question and responses
(sample each response with 0.5 probability). 
\emph{Skill-set-based replacement} is almost the same as the original skill-based replacement, but the candidates of the questions to be replaced are chosen as ones with exactly same set of skills are associated, not only have common skills ($S = S^{\rep}$).
The results in Table \ref{tab:rep_skill} show that the performances of the question random replacements depends on the nature of dataset.
It shows similar performance with skill-based replacement on ASSISTmentsChall and EdNet-KT1 datasets, but only give a minor gain or even dropped the performance on other datasets. 
However, applying interaction-random replacement significantly hurt performances over all datasets, e.g. the AUC is decreased from 86.43\% to 84.50\% on STATICS2011 dataset.  
This demonstrates the importance of fixing responses of the interactions  for consistency regularization.
At last, skill-set-based replacement works similar or even worse than the original skill-based replacement. 
Note that each question of the STATICS2011 dataset has single skill attached to, so the performance of skill-based and skill-set-based replacement coincide on the dataset.

\textbf{Comparison with other regularization methods in KT} We also compare our regularization scheme with previous works: DKT+ \cite{yeung2018addressing} and qDKT \cite{sonkar2020qdkt}.
DKT+ uses two types of regularization losses: \emph{reconstruction loss} and \emph{waviness loss}.
Reconstruction loss enable a model to recover the current interaction's label, and waviness loss make model's prediction to be consistent over all timesteps. 
% These losses are defined as follows:
% \begin{align}
%     \mathcal{L}_{r} &= \mathbb{E}_{t \in [T-1]}[\ell(p_{t, j(t)}, R_{t})], \\
%     \mathcal{L}_{w_{1}} &= \mathbb{E}_{t\in [T-1], j\in [\mathbf{Q}]}[|p_{t+1, j} - p_{t, j}|], \\
%     \mathcal{L}_{w_{2}} &= \mathbb{E}_{t\in [T-1], j \in [\mathbf{Q}]}[(p_{t+1, j} - p_{t, j})^{2}],
% \end{align}
% where $\ell$ is a BCE loss, $p_{t, j}$ is the predicted correctness probability of question $q_j = q_{j(t)}$ at step $t$, and $\mathbf{Q}$ is the total number of questions.
% qDKT that uses the following Laplacian loss which regularizes the variance of predicted correctness probabilities for questions that fall under the same skill: 
qDKT uses the Laplacian loss that regularizes the variance of predicted correctness probabilities for questions that fall under the same skill, which is similar to the variation $\mathcal{L}_{\regu\shyp\rep\_\mathrm{ro}}$ of the consistency loss.
% \begin{align}
%     \mathcal{L}_{\mathrm{Laplacian}} = \mathbb{E}_{(q_i, q_j) \in \mathcal{Q}\times\mathcal{Q}} [\bm{1}(i, j)(p_{i} - p_{j})^{2}]
% \end{align}
% where $\mathcal{Q}$ is the set of all questions, $p_i, p_j$ are the model's predicted correctness probabilities for the questions $q_i, q_j$, and $\bm{1}(i, j)$ is 1 if $q_i, q_j$ have common skills attached, otherwise 0. 
% It is similar to our variation of consistency regularization $\mathcal{L}_{\regu\shyp\rep\_\mathrm{ro}}$  that only compares replaced interactions' outputs, but it does not replace questions and it compares all questions (with same skills) at once. 
We explain these losses in detail in the Appendix.

Results in Table \ref{tab:comparison} shows that our regularization approach yields the largest performance gain over all benchmarks compared to other methods. 
In some cases, using DKT+ or qDKT even harm the performances, while consistency and monotonocity regularization yields substantial performance gain over all datasets.
% \vspace{-0.1in}
% \vspace{-0.05in}
\section{Conclusion}

We propose simple augmentation strategies with corresponding constraint regularization losses for KT and show their efficacy. 
%As in computer vision, it is natural to consider compositions of the augmentations, such as replacing interactions after random deletion. 
%Suitable regularization losses for each composition might increase model's performance further, as diversity of augmented sequences increases.
% and comparing it with real sequences. 
We only considered the most basic features of interactions, question and response correctness, and other features like elapsed time or question texts enables us to exploit diverse augmentation strategies if available.
Furthermore, exploring applicability of our idea on other AIEd tasks (dropout prediction or at-risk student prediction) is another interesting future direction.
%possible direction of follow-up study. 
% At last, it may possible tconsistency and monotonicity learning framework may 
% If other features such as elapsed time or question texts are available, it may possible to utilize the additional 
% Also, considering other interaction features like elapsed times 
% At last, 
% Compositions of augmentations, other features, other tasks

% \subsubsection*{Acknowledgments}
% Thank you
% Use unnumbered third level headings for the acknowledgments. All
% acknowledgments, including those to funding agencies, go at the end of the paper.

\newpage
\bibliographystyle{named}
\bibliography{ijcai21}

\end{document}

\maketitle

\begin{table*}[!htp]\centering
\resizebox{1.5\columnwidth}{!}{
\scriptsize
\begin{tabular}{lcccccccc}\toprule
name & logs & students & questions & skills & avg. length & avg. correctness \\ \midrule
% ASSIST2009 & & & & \\
ASSIST2015 & 683801 & 19840& 100 & - & 34.47 &   0.73 \\
ASSISTChall & 942816 & 1709 & 3162 & 102 & 551.68  & 0.37 \\ 
STATICS2011 & 261937 & 333 & 1224 & 81 & 786.60  & 0.72 \\ 
% JunYi & & & & \\
% KDDCup2010 & & & & \\
EdNet-KT1 & 2051701 & 6000 & 14419 & 188 & 341.95 & 0.63 \\
\bottomrule
\end{tabular}
}
\label{tab:datastat}
\caption{Dataset statistics.}
\end{table*}

\section{Dataset}

\subsection{Dataset statistics and pre-processing}

% We demonstrated the effectiveness of the proposed method on 4 widely used benchmark datasets: ASSISTments2015, ASSISTmentsChall, STATICS2011, and EdNet-KT1. 
\textbf{ASSISTments} datasets are the most widely used benchmark for Knowledge Tracing, which is provided by ASSISTments online tutoring platform\footnote{https://new.assistments.org/}. There are several versions of dataset depend on the years they collected, and we used ASSISTments2015\footnote{https://sites.google.com/site/assistmentsdata/home/2015-assistments-skill-builder-data} and ASSISTmentsChall\footnote{https://sites.google.com/view/assistmentsdatamining}. 
ASSISTmentsChall dataset is provided by the 2017 ASSISTments data mining competition. 
For ASSISTments2015 dataset, we filtered out the logs with \texttt{CORRECTS} not in $\{0,1\}$. 
Note that ASSISTments2015 dataset only provides question and no corresponding skills.

\textbf{STATICS2011} consists of the interaction logs from an engineering statics course, which is available on the PSLC datashop\footnote{https://pslcdatashop.web.cmu.edu/DatasetInfo?datasetId=507}. A concatenation of a \texttt{problem name} and \texttt{step name} is used as a question id, and the values in the column \texttt{KC (F2011)} are regarded as skills attached to each question. 

\textbf{EdNet-KT1} is the largest publicly available interaction dataset consists of TOEIC (Test of English for Interational Communication) problem solving logs collected by \emph{Santa}\footnote{https://aitutorsanta.com/}.
We reduce the size of the EdNet-KT1 dataset by sampling 6000 users among 600K users.
Detailed statistics and pre-processing methods for these datasets are described in Appendix. 
With the exception of the EdNet-KT1 dataset, we used 80\% of the students as a training set and the remaining 20\% as a test set. 
Among 600K students, we filtered out whose interaction length is in $[100, 1000]$, and randomly sampled 6000 users, where 5000 users for training and 1000 users for test. 

Detailted dataset statistics are given in the Table 1 below. 
% \begin{itemize}
%     \item \textbf{ASSISTments}: 
%     For ASSISTments2015\footnote{https://sites.google.com/site/assistmentsdata/home/2015-assistments-skill-builder-data} dataset, we filtered out the logs with \texttt{CORRECTS} not in $\{0,1\}$. 
%     Note that ASSISTments2015 dataset only provides question and no corresponding skills.
%     \item \textbf{STATICS2011} 
%     A concatenation of a \texttt{problem name} and \texttt{step name} is used as a question id, and the values in the column \texttt{KC (F2011)} are regarded as skills attached to each question. 

%     % \item \textbf{Junyi Academy} is a dataset collected from Junyi Academy\footnote{https://www.junyiacademy.org}, the online learning platform in China.
%     % It was first introduced in \cite{chang2015modeling} and available on the PSLC datashop\footnote{https://pslcdatashop.web.cmu.edu/DatasetInfo?datasetId=1275}. 

%     % \item \textbf{KDDCup}

%     \item \textbf{EdNet-KT1}
%     Among 600K students, we filtered out whose interaction length is in $[100, 1000]$, and randomly sampled 6000 users, where 5000 users for training and 1000 users for test. 

% \end{itemize}

\subsection{Monotonicity nature of datasets}
\label{sec:mono_data}

We perform data analysis to explore monotonicity nature of datasets, i.e. a property that students are more likely to answer correctly if they did the same more in the past. 
% We first fix response among correct and incorrect, and check whether 
For each interaction of each student, we see the distribution of past interactions' correctness rate.
Formally, for given interaction sequences $(I_1, \dots, I_T)$ with $I_t = (Q_t, R_t)$ and each $2\leq t\leq T$, we compare the distributions of past interactions' correctness rate
$$
\text{correctness\_rate}_{<t} = \frac{1}{t-1} \sum_{\tau=1}^{t-1} \bm{1}_{R_{\tau} = 1}
$$
where $\bm{1}_{R_{\tau}=1}$ is an indicator function which is $1$ (resp. $0$) when $R_{\tau} = 1$ (resp. $R_{\tau} = 0$). 
We compare the distributions of $\text{correctness\_rate}_{<t}$ over all interactions with $R_{t} = 1$ and $R_{t} = 0$ separately, and the results are shown in Figure 1 of the main text.
We can see that the distributions of previous correctness rates of interactions with  correct response lean more to the right than ones of interactions with incorrect response.
This shows the positive correlation between previous correctness rate and the current response correctness, and it also explains why monotonicity regularization actually improve prediction performances of knowledge tracing models.

\begin{table}[t]
% \vspace{-0.1in}
\centering
%\resizebox{\columnwidth}{!}
{
\small
\begin{tabular}{cccc}
\toprule
dataset & target response & vanilla & regularized \\
\midrule
\multirow{2}{*}{ASSISTChall} & correct & 0.01028 & 0.00027\\
& incorrect & 0.01713 & 0.00039\\
\midrule
\multirow{2}{*}{STATICS2011} & correct & 0.00618 & 0.00049 \\
& incorrect & 0.01748 & 0.00093 \\
\midrule
\multirow{2}{*}{EdNet-KT1} & correct & 0.00422 & 0.00091 \\
& incorrect & 0.00535 & 0.00116 \\
\bottomrule
\end{tabular}
}
\caption{Comparison of the average consistency loss for correctly and incorrectly predicted responses of the DKT model.}

%For each dataset, upper row represents the results with only data augmentation, and the lower row represents the results with augmentation and regularization losses. }
% Training the model only with data augmentation is not enough for the meaningful improvement in AUC, but including consistency and monotonicity regularization losses boost up the performances.  
\label{tab:cons_loss}
% \vspace{-0.15in}
\end{table}

\section{Model}
\subsection{Model's predictions and consistency regularization losses}
\label{sec:test_cons_loss}

Instead of analyzing consistency nature of datasets directly, 
we compare the test consistency loss for correctly and incorrectly predicted responses separately, with the DKT model on ASSISTmentsChall, STATICS2011, and EdNet-KT1 datasets.
Table \ref{tab:cons_loss} shows the average consistency loss for correctly and incorrectly predicted responses, with the vanilla DKT model and the model trained with consistency regularization losses.
When we compute the test consistency loss, we replaced each (previous) interaction's questions to another questions with overlapping skills with $\alpha_{\rep} = 0.3$ probability. 
For all models, the average loss for the correctly predicted responses are lower than the incorrectly predicted responses.
This verifies that smaller consistency loss actually improves prediction accuracy.

\subsection{Overfitting phenomena}
\begin{figure*}[t]
\vspace{-0.15in}
\centering
\begin{tikzpicture}[every mark/.append style={mark size=1pt}]
\begin{axis}[
    title style={at={(0.5,0)},anchor=north,yshift=-15.0},
    title={ASSISTments2015},
    % width=0.30\linewidth,
    width=0.6\columnwidth,
    at={(-520,0)},
    xmin=0, xmax=4100,
    ymin=67.5, ymax=73,
    xtick={1000, 2000, 3000, 4000},
    xticklabels={1000, 2000, 3000, 4000},
    ytick={66, 68, 70, 72, 74},
    legend pos=south east,
    ymajorgrids=true,
    grid style=dashed,
    legend style={nodes={scale=0.6}},
    yticklabel style = {font=\tiny,xshift=0.5ex},
    xticklabel style = {font=\tiny,yshift=0.5ex}
    % legend image post style={mark=*}
]
\addplot[
    color=red,
    mark=square,
    ]
    coordinates {
    (100,54.0)(200,61.09)(300,64.5)(400,65.95)(500,67.15)(600,68.08)(700,68.95)(800,69.93)(900,70.88)(1000,71.45)(1100,71.73)(1200,71.84)(1300,71.85)(1400,71.94)(1500,71.95)(1600,71.91)(1700,72.13)(1800,71.97)(1900,72.03)(2000,72.14)(2100,72.03)(2200,72.01)(2300,72.1)(2400,72.1)(2500,71.97)(2600,72.12)(2700,72.09)(2800,72.19)(2900,72.18)(3000,72.07)(3100,72.2)(3200,72.06)(3300,72.0)(3400,71.82)(3500,72.07)(3600,71.87)(3700,71.96)(3800,72.06)(3900,72.06)(4000,72.12)
    };
\addplot[
    color=blue,
    mark=asterisk,
    ]
    coordinates {
    (100,53.82)(200,62.67)(300,65.96)(400,67.62)(500,69.48)(600,70.7)(700,71.19)(800,71.64)(900,71.69)(1000,71.51)(1100,71.66)(1200,71.76)(1300,71.51)(1400,71.43)(1500,71.37)(1600,71.25)(1700,71.59)(1800,71.42)(1900,71.1)(2000,71.09)(2100,71.16)(2200,70.75)(2300,70.89)(2400,70.89)(2500,70.33)(2600,70.51)(2700,70.41)(2800,70.54)(2900,70.11)(3000,69.96)(3100,69.67)(3200,70.0)(3300,69.77)(3400,69.48)(3500,69.63)(3600,69.52)(3700,69.58)(3800,69.39)(3900,69.2)(4000,69.38)
    };
% \legend{w/ augmentation, w/o augmentation}
% \legend{no augmentation, insertion, deletion}
\end{axis}

% assistchall
\begin{axis}[
    title style={at={(0.5,0)},anchor=north,yshift=-15.0},
    title={ASSISTmentsChall},
    % width=0.30\linewidth,
    width=0.6\columnwidth,
    at={(0,0)},
    xmin=0, xmax=4100,
    ymin=60, ymax=80,
    xtick={1000, 2000, 3000, 4000},
    xticklabels={1000, 2000, 3000, 4000},
    ytick={60,64,68,72,76,80},
    legend pos=south east,
    ymajorgrids=true,
    grid style=dashed,
    legend style={nodes={scale=0.6}},
    yticklabel style = {font=\tiny,xshift=0.5ex},
    xticklabel style = {font=\tiny,yshift=0.5ex}
    % legend image post style={mark=*}
]
% rep + ins + del
\addplot[
    color=red,
    mark=square,
    ]
    coordinates {
    (100,54.89)(200,63.17)(300,67.41)(400,68.12)(500,68.88)(600,70.38)(700,71.71)(800,72.38)(900,72.93)(1000,73.39)(1100,73.72)(1200,74.34)(1300,75.06)(1400,75.73)(1500,76.46)(1600,77.09)(1700,77.34)(1800,77.69)(1900,77.94)(2000,78.13)(2100,78.3)(2200,78.45)(2300,78.53)(2400,78.59)(2500,78.76)(2600,78.79)(2700,78.85)(2800,78.85)(2900,78.89)(3000,78.93)(3100,78.95)(3200,79.02)(3300,78.91)(3400,78.91)(3500,78.9)(3600,78.99)(3700,78.9)(3800,78.88)(3900,78.8)(4000,78.93)
    };
\addplot[
    color=blue,
    mark=asterisk,
    ]
    coordinates {
    (100,56.61)(200,64.84)(300,68.32)(400,70.18)(500,71.56)(600,72.51)(700,73.32)(800,73.86)(900,74.22)(1000,74.37)(1100,74.44)(1200,74.4)(1300,74.39)(1400,74.43)(1500,74.45)(1600,74.22)(1700,74.05)(1800,74.01)(1900,73.99)(2000,73.8)(2100,73.78)(2200,73.57)(2300,73.43)(2400,73.34)(2500,73.14)(2600,73.05)(2700,72.84)(2800,72.89)(2900,72.69)(3000,72.52)(3100,72.64)(3200,72.45)(3300,72.28)(3400,72.27)(3500,72.18)(3600,72.12)(3700,71.92)(3800,71.94)(3900,71.75)(4000,71.92)
    };

% \legend{no augmentation, replacement, insertion, deletion}
% \legend{w/ augmentation, w/o augmentation}
\end{axis}

% statics
\begin{axis}[
    % width=0.30\linewidth,
    width=0.6\columnwidth,
    at={(260, 0)},
    title style={at={(0.5,0)},anchor=north,yshift=-15.0},
    title={STATICS2011},
    xmin=0, xmax=2100,
    ymin=76, ymax=88,
    xtick={500, 1000, 1500, 2000},
    xticklabels={500, 1000, 1500, 2000},
    ytick={76, 78, 80, 82, 84, 86, 88},
    legend pos=south east,
    ymajorgrids=true,
    grid style=dashed,
    legend style={nodes={scale=0.6}},
    yticklabel style = {font=\tiny,xshift=0.5ex},
    xticklabel style = {font=\tiny,yshift=0.5ex}
    % legend image post style={mark=*}
]
% rep + ins + del
\addplot[
    color=red,
    mark=square,
    ]
    coordinates {
    (100,62.13)(200,77.09)(300,77.2)(400,80.15)(500,82.05)(600,83.59)(700,84.39)(800,84.96)(900,85.43)(1000,85.84)(1100,86.24)(1200,86.45)(1300,86.65)(1400,86.69)(1500,86.84)(1600,86.85)(1700,86.89)(1800,86.88)(1900,86.92)(2000,86.91)
    };
\addplot[
    color=blue,
    mark=asterisk,
    ]
    coordinates {
    (100,63.41)(200,77.67)(300,81.08)(400,84.78)(500,85.98)(600,86.56)(700,86.57)(800,86.51)(900,86.36)(1000,85.99)(1100,85.86)(1200,85.6)(1300,85.36)(1400,85.08)(1500,84.95)(1600,84.71)(1700,84.56)(1800,84.28)(1900,84.11)(2000,83.84)
    };

% \legend{no augmentation, replacement, insertion, deletion}
% \legend{w/ augmentation, w/o augmentation}
\end{axis}

% ednet

\begin{axis}[
    title style={at={(0.5,0)},anchor=north,yshift=-15.0},
    % width=0.30\linewidth,
    width=0.6\columnwidth,
    at={(770, 0)},
    title={EdNet-KT1},
    xmin=0, xmax=3100,
    ymin=67, ymax=74,
    xtick={1000, 2000, 3000},
    xticklabels={1000, 2000, 3000},
    ytick={68, 70, 72, 74},
    legend pos=south east,
    ymajorgrids=true,
    grid style=dashed,
    legend style={nodes={scale=0.5}},
    yticklabel style = {font=\tiny,xshift=0.5ex},
    xticklabel style = {font=\tiny,yshift=0.5ex}
    % legend image post style={mark=*}
]
% rep + ins + del
\addplot[
    color=red,
    mark=square,
    ]
    coordinates {
    (100,51.25)(200,60.09)(300,68.18)(400,68.75)(500,68.9)(600,68.96)(700,69.09)(800,69.81)(900,70.63)(1000,71.37)(1100,71.74)(1200,72.13)(1300,72.4)(1400,72.53)(1500,72.64)(1600,72.77)(1700,72.91)(1800,72.97)(1900,73.01)(2000,73.07)(2100,73.06)(2200,73.09)(2300,73.08)(2400,73.05)(2500,72.93)(2600,72.96)(2700,72.92)(2800,72.82)(2900,72.81)(3000,72.81)
    };
\addplot[
    color=blue,
    mark=asterisk,
    ]
    coordinates {
    (100,51.27)(200,63.34)(300,68.44)(400,69.05)(500,69.18)(600,69.21)(700,69.2)(800,69.56)(900,69.91)(1000,70.26)(1100,70.59)(1200,70.68)(1300,71.06)(1400,71.17)(1500,71.33)(1600,71.53)(1700,71.51)(1800,71.56)(1900,71.52)(2000,71.4)(2100,71.26)(2200,71.12)(2300,71.0)(2400,70.85)(2500,70.66)(2600,70.44)(2700,70.3)(2800,70.04)(2900,69.99)(3000,69.72)
    };
% \legend{w/ augmentation, w/o augmentation}
\end{axis}

\end{tikzpicture}
\caption{Validation AUCs of DKT model and it's regularized version. Red curve (resp. blue curve) represents validation aucs of regularized DKT (resp. vanilla DKT) model. We can see that the vanilla DKT quickly overfits and it's validation AUC starts to decrease early, while regularization make the model to less overfit and improve model's performance. }
\label{fig:val_auc}
\vspace{-0.1in}
\end{figure*}
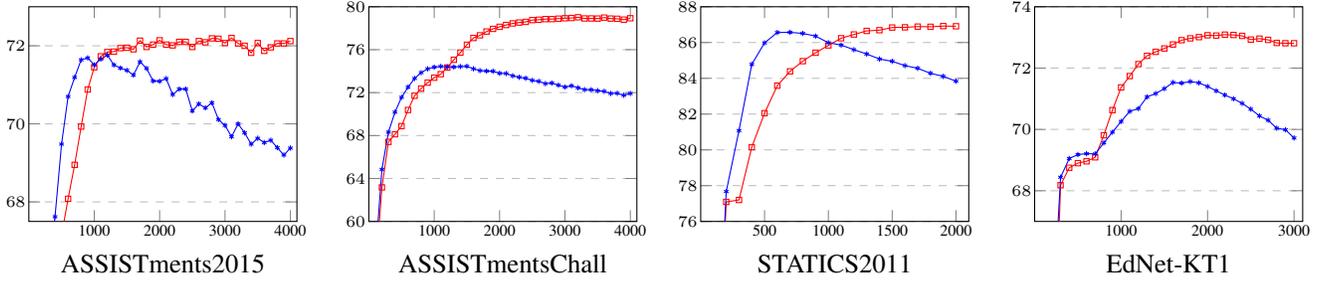

In Figure \ref{fig:val_auc}, we plot the graph of validaion AUCs of vanilla DKT model and regularized DKT model. Red curve (resp. blue curve) represents the AUCs of regularized DKT model (resp. vanilla DKT model). 
We can observe that vanilla DKT model quickly overfits to training set, which makes the validation AUC decrease. 
However, when we train the model with our regularization losses (with suitable hyperparameters), the model overfits less and it's performance also increases.

\subsection{Hyperparameters}

\subsubsection{Hyperparameters for the main table}
Table \ref{tab:hyper_main}  describes detailed hyperparameters for each augmentation and model that are used for the main results (Table 1 of the main text). 
Each entry represents a tuple of augmentation probability ($\alpha_{\aug}$) and a weight for constraint loss ($\lambda_{\regu\shyp\aug}$), which shows the best performances among $\alpha_{\aug} \in \{0.1, 0.3, 0.5\}$ and $\lambda_{\regu\shyp\aug}\in \{ 1, 10, 50, 100\}$. 
Each entry represents $(\alpha_{\aug}, \lambda_{\regu\shyp\aug})$ for each augmentation. 
We use $\lambda_{\aug}=1$ for all experiments with augmentations, except for the DKT model on STATICS2011 dataset with incorrect insertion augmentation ($\lambda_{\incor\_\ins}=0$).

To see the effect of augmentation probabilities and regularization loss weights, we perform grid search over $\alpha_{\aug} \in \{0.1, 0.3, 0.5\}$ and $\lambda_{\regu\shyp\aug} \in \{1, 10, 50, 100\}$ with DKT model, and the AUC results are shown as heatmaps in Figure 1. 

% \textcolor{red}{Need to say something about this}

\begin{table}[t]
% \vspace{-0.1in}
\centering
%\resizebox{\columnwidth}{!}
{
\small
\begin{tabular}{@{\extracolsep{4pt}}cccccc@{}}
\toprule
\multirow{2}{*}{dataset} &  \multicolumn{3}{c}{DKT+} & \multicolumn{1}{c}{qDKT} \\ 
\cmidrule{2-4} \cmidrule{5-5}
& $\lambda_{r}$ & $\lambda_{w_{1}}$ & $\lambda_{w_{2}}$ & $\lambda$ \\
\midrule
ASSIST15 & 0.05 & 0.03 & 3.0 & - \\
\midrule
ASSISTChall & 0.1 & 0.3 & 3.0 & 0.1 \\
\midrule
STATICS2011 & 0.2 & 1.0 & 30.0 & 0.5 \\
\midrule
EdNet-KT1 & 0.1 & 0.1 & 10.0 & 0.01 \\
\bottomrule
\end{tabular}
}
\caption{Hyperparamters for DKT+ and qDKT.}

%For each dataset, upper row represents the results with only data augmentation, and the lower row represents the results with augmentation and regularization losses. }
% Training the model only with data augmentation is not enough for the meaningful improvement in AUC, but including consistency and monotonicity regularization losses boost up the performances.  
\label{tab:dktplus}
% \vspace{-0.15in}
\end{table}

\subsubsection{Losses and hyperparameters for DKT+ and qDKT}

DKT+ uses two types of regularization losses: \emph{reconstruction loss} and \emph{waviness loss}.
Reconstruction loss enable a model to recover the current interaction's label, and waviness loss make model's prediction to be consistent over all timesteps. 
These losses are defined as follows:
\begin{align}
    \mathcal{L}_{r} &= \mathbb{E}_{t \in [T-1]}[\ell(p_{t, j(t)}, R_{t})], \\
    \mathcal{L}_{w_{1}} &= \mathbb{E}_{t\in [T-1], j\in [\mathbf{Q}]}[|p_{t+1, j} - p_{t, j}|], \\
    \mathcal{L}_{w_{2}} &= \mathbb{E}_{t\in [T-1], j \in [\mathbf{Q}]}[(p_{t+1, j} - p_{t, j})^{2}],
\end{align}
where $\ell$ is a BCE loss, $p_{t, j}$ is the predicted correctness probability of question $q_j = q_{j(t)}$ at step $t$, and $\mathbf{Q}$ is the total number of questions. 
After that, DKT+ is trained with a new loss function
$$
\mathcal{L}_{\mathrm{DKT}+} = \mathcal{L}_{\mathrm{KT}} + \lambda_{r} \mathcal{L}_{r} + \lambda_{w_{1}} \mathcal{L}_{w_{1}} + \lambda_{w_{2}} \mathcal{L}_{w_{2}}
$$
with suitable choice of scaling constants $\lambda_{r}, \lambda_{w_{1}}, \lambda_{w_{2}}$.

qDKT that uses the following Laplacian loss which regularizes the variance of predicted correctness probabilities for questions that fall under the same skill: 
\begin{align}
    \mathcal{L}_{\mathrm{Laplacian}} = \mathbb{E}_{(q_i, q_j) \in \mathcal{Q}\times\mathcal{Q}} [\bm{1}(i, j)(p_{i} - p_{j})^{2}]
\end{align}
where $\mathcal{Q}$ is the set of all questions, $p_i, p_j$ are the model's predicted correctness probabilities for the questions $q_i, q_j$, and $\bm{1}(i, j)$ is 1 if $q_i, q_j$ have common skills attached, otherwise 0. 
It is similar to our variation of consistency regularization $\mathcal{L}_{\regu\shyp\rep\_\mathrm{ro}}$  that only compares replaced interactions' outputs, but it does not replace questions and it compares all questions (with same skills) at once. 
Then qDKT is trained with a new loss function
$$
\mathcal{L}_{\mathrm{qDKT}} = \mathcal{L}_{\mathrm{KT}} + \lambda \mathcal{L}_{\mathrm{Laplacian}}
$$
with suitable choice of a scaling constant $\lambda$.

Table \ref{tab:dktplus} describes the hyperparameters, i.e. scaling constants for each loss (reconstruction loss, waviness loss, and laplacian loss).
When we train DKT+, the best combinations of hyperparameters that is reported in the original paper are used for ASSISTments 2015, ASSISTmentsChall, and STATICS2011 dataset, and we search over the range suggested in the paper and choose the combination with best result for EdNet-KT1.
For qDKT, since the coefficient $\lambda$ for the Laplacian loss is not given in the original paper, we choose $\lambda$ among $\{0.01, 0.05, 0.1, 0.5, 1,10, 50, 100\}$, and report the best result.

% NEW
\begin{table*}[!htp]\centering
\resizebox{1.9\columnwidth}{!}{
\normalsize
\begin{tabular}{@{\extracolsep{4pt}}ccccccccccc@{}}\toprule
\multirow{2}{*}{dataset} & \multirow{2}{*}{model} & \multicolumn{4}{c}{insertion  + deletion} & \multicolumn{5}{c}{insertion + deletion + replacement} \\ 
\cmidrule{3-6} \cmidrule{7-11}
& & cor\_ins & incor\_ins & cor\_del & incor\_del & cor\_ins & incor\_ins & cor\_del & incor\_del & rep\\
\midrule
% ASSIST2009 & & & & & & \\ 
\multirow{3}{*}{ASSIST2015} & DKT & (0, 0) & (0, 0) & (0.3, 100) & (0, 0) & (0.3, 100) & (0, 0) & (0, 0) & (0, 0) & (0.1, 10) \\
& DKVMN & (0.5, 100) & (0, 0) & (0, 0) & (0, 0)  & (0.5, 100) & (0, 0) & (0, 0) & (0, 0) & (0.3, 1) \\
% & SAKT & \\
& SAINT & (0, 0) & (0.5, 10) & (0, 0) & (0, 0) & (0, 0) & (0.5, 10) & (0, 0) & (0, 0) & (0.3, 1)\\ \midrule
\multirow{3}{*}{ASSISTChall} & DKT & (0.5, 100) & (0, 0) & (0, 0) & (0, 0) & (0.5, 1) & (0, 0) & (0, 0) & (0, 0) & (0.3, 100) \\
& DKVMN & (0.5, 1) & (0, 0) & (0, 0) & (0, 0) & (0.5, 1) & (0, 0) & (0, 0) & (0, 0) & (0.5, 100)  \\
% & SAKT & \\
& SAINT & (0, 0) & (0, 0) & (0.3, 1) & (0, 0) & (0, 0) & (0.3, 1) & (0.3, 1) & (0, 0) & (0.3, 100)\\ \midrule
\multirow{3}{*}{STATICS2011} & DKT & (0, 0) & (0.5, 10) & (0, 0) & (0, 0) & (0, 0) & (0, 0) & (0, 0) & (0, 0) & (0.3, 100) \\
& DKVMN & (0, 0) & (0, 0) & (0.3, 10) & (0, 0) & (0, 0) &  (0, 0) & (0.3, 1) & (0, 0) & (0.3, 10) \\
% & SAKT & \\
& SAINT & (0, 0) & (0.5, 1) & (0, 0) & (0.5, 1) & (0, 0) & (0.5, 1) & (0, 0) & (0.5, 1) & (0.3, 100) \\ \midrule
% JunYi & & & & \\
% KDDCup2010 & & &  & & & \\
\multirow{3}{*}{EdNet-KT1} & DKT &(0, 0)& (0, 0) & (0.3, 50) & (0, 0) & (0, 0) & (0.3, 1) & (0.3, 1) & (0, 0) & (0.1, 100)  \\
& DKVMN & (0, 0) & (0.5, 1) & (0, 0) & (0, 0) & (0, 0) & (0.5, 1) & (0, 0) & (0, 0) & (0.1, 1) \\
% & SAKT & \\
& SAINT & (0, 0) & (0.3, 50) & (0, 0) & (0, 0) & (0, 0) & (0.3, 50) & (0, 0) & (0, 0) & (0.5, 1) \\ 
\bottomrule
\end{tabular}
}
\caption{Hyperparameters for Table 1 of the main text.}
\label{tab:hyper_main}
\end{table*}

\begin{figure*}[htp]
\centering
\begin{subfigure}{1.2\textwidth}
    \hspace*{-2cm}
    \includegraphics[clip,width=\columnwidth]{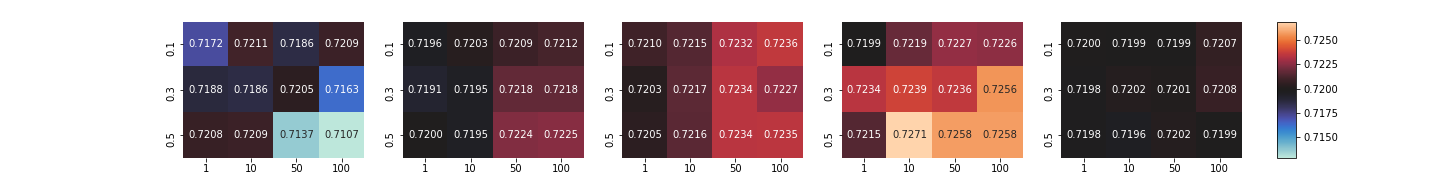}%
    \caption{ASSISTments2015}
\end{subfigure}
\begin{subfigure}{1.2\textwidth}
    \hspace*{-2cm}
    \includegraphics[clip,width=\columnwidth]{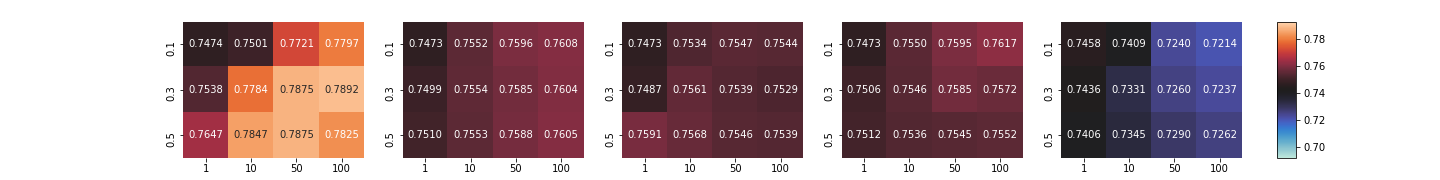}%
    \caption{ASSISTmentsChall}
\end{subfigure}
\begin{subfigure}{1.2\textwidth}
    \hspace*{-2cm}
    \includegraphics[clip,width=\columnwidth]{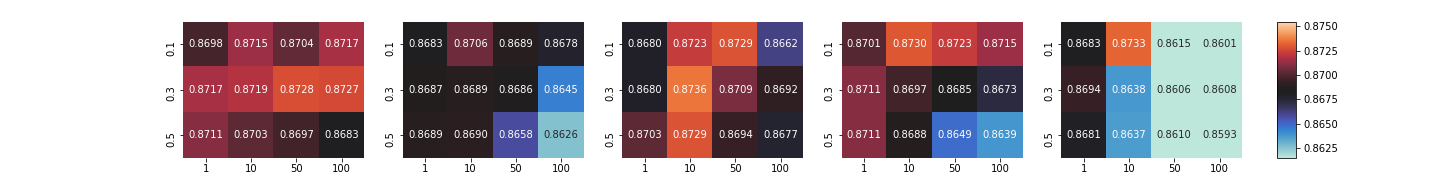}%
    \caption{STATICS2011}
\end{subfigure}
\begin{subfigure}{1.2\textwidth}
    \hspace*{-2cm}
    \includegraphics[clip,width=\columnwidth]{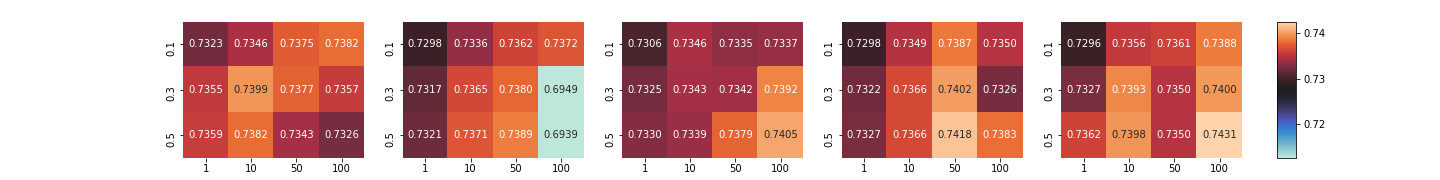}%
    \caption{EdNet-KT1}
\end{subfigure}
\label{fig:hp_grid_all}
\caption{
Test AUCs of the DKT model for each augmentation and corresponding regularization with different augmentation probabilities ($\alpha_{\aug})$) and regularization loss weights ($\lambda_{\regu-\aug}$). 
The hyperparameters are searched over $\alpha_{\aug} \in \{0.1, 0.3, 0.5\}$ and $\lambda_{\regu-\aug}\in \{1, 10, 50, 100\}$.
For each dataset, each column represents results with replacement, correct insertion, incorrect insertion, correct deletion, and incorrect deletion, from left to right.
We set $\lambda_{\aug} = 1$ for all cases.
We use question-random replacement for ASSISTments2015 dataset.
}

\end{figure*}

% \bibliographystyle{named}
% \bibliography{ijcai21}